\documentclass[lettersize,journal]{IEEEtran}
\usepackage{amsmath,amsfonts}
\usepackage{algorithmic}
\usepackage{algorithm}
\usepackage{array}
\usepackage[caption=false,font=normalsize,labelfont=sf,textfont=sf]{subfig}
\usepackage{textcomp}
\usepackage{stfloats}
\usepackage{url}
\usepackage{verbatim}
\usepackage{graphicx}
\usepackage{cite}
\usepackage{booktabs}
\usepackage{multirow}
\usepackage[table]{xcolor}
\usepackage{placeins}

\begin{document}

\title{Enhancing Targeted Adversarial Attacks on Large Vision-Language Models via Intermediate Projector}

\author{Yiming Cao, Yanjie Li, Kaisheng Liang, Bin Xiao,~\IEEEmembership{Fellow,~IEEE}
\thanks{Y. Cao, Y.Li, K.Liang and B.Xiao are with the Department of Computing, The Hong Kong Polytechnic University (Email: \{yiming.cao, yanjie.li, kaisheng.liang\}@connect.polyu.hk; b.xiao@polyu.edu.hk).}
}

\maketitle

\begin{abstract}
The growing deployment of Large Vision-Language Models (VLMs) raises safety concerns, as adversaries may exploit model vulnerabilities to induce harmful outputs, with targeted black-box adversarial attacks posing a particularly severe threat.
However, existing methods primarily maximize encoder-level global similarity, which lacks the granularity for stealthy and practical fine-grained attacks, where only specific target should be altered (e.g., modifying a car while preserving its background). Moreover, they largely neglect the projector, a key semantic bridge in VLMs for multimodal alignment.
To address these limitations, we propose a novel black-box targeted attack framework that leverages the projector. Specifically, we utilize the widely adopted Querying Transformer (Q-Former) which transforms global image embeddings into fine-grained query outputs, to enhance attack effectiveness and granularity.
For standard global targeted attack scenarios, we propose the Intermediate Projector Guided Attack (IPGA), which aligns Q-Former’s fine-grained query outputs with the target to enhance attack strength and exploits the intermediate pretrained Q-Former that is not fine-tuned for any specific Large Language Model (LLM) to improve attack transferability.
For fine-grained attack scenarios, we augment IPGA with the Residual Query Alignment (RQA) module, which preserves unrelated content by constraining non-target query outputs to enhance attack granularity.
Extensive experiments demonstrate that IPGA significantly outperforms baselines in global targeted attacks, and IPGA with RQA (IPGA-R) attains superior success rates and unrelated content preservation over baselines in fine-grained attacks. Our method also transfers effectively to commercial VLMs such as Google Gemini and OpenAI GPT. 
\end{abstract}


\section{Introduction}
\IEEEPARstart{L}{arge} Vision-Language Models (VLMs) typically integrate Large Language Models (LLMs) with pretrained visual encoders via a projection module, enabling joint visual-textual reasoning~\cite{li2023blip, liu2024visual, lu2024deepseek}. They achieve strong performance on a wide range of multimodal tasks from answering general questions to task planning for AI agents and robots~\cite{driess2023palm, luo2024vision}. However, recent studies have revealed that VLMs are highly vulnerable to image adversarial attacks~\cite{wang2024transferable, cui2024robustness}, which can lead to severe consequences in downstream applications~\cite{wu2024dissecting}. As VLMs become more accessible to the public, assessing the robustness of VLMs against such attacks is crucial prior to real-world deployment~\cite{zhao2024evaluating, guo2024efficient}.

While early evaluations of VLM robustness have focused on untargeted attacks~\cite{wang2024transferable, cui2024robustness}, targeted attacks pose a greater risk by inducing specific, potentially harmful outputs~\cite{zhao2024evaluating}. 
The targeted attack becomes more challenging and practical under the black-box setting, where the attacker has no access to the victim model's internal information~\cite{guo2024efficient}. Therefore, we investigate the black-box targeted attacks against VLMs.
Existing methods typically optimize global image embeddings (e.g., the [CLS] token embedding from pretrained encoders like CLIP~\cite{radford2021learning}) to align with target text or reference image~\cite{zhao2024evaluating, xie2024chain}, limiting perturbations to global image content modifications.
However, these coarse global perturbations are insufficient for fine-grained targeted attacks that modify only the target while preserving the surrounding unrelated content. Such fine-grained control over adversarial perturbations is crucial for achieving attack stealth and practical utility in real-world applications that require detailed visual understanding like autonomous driving~\cite{zheng2024physical}. For instance, modifying a target vehicle without affecting its background is necessary to maintain plausibility; without such granularity, perturbations risk spilling into unrelated regions, exposing the attack and undermining its effect.
Furthermore, existing methods largely neglect the projector, a critical semantic bridge in VLMs that connects the visual encoder to the LLM. Trained on large-scale multimodal data, the projector captures multimodal alignment patterns beyond the encoder’s capacity~\cite{li2023blip}. Ignoring the projector limits the attack’s ability to disrupt the full vision-language alignment pipeline in VLM, consequently reducing attack effectiveness.

\begin{figure*}[!t]
    \centering
    \includegraphics[width=1\linewidth]{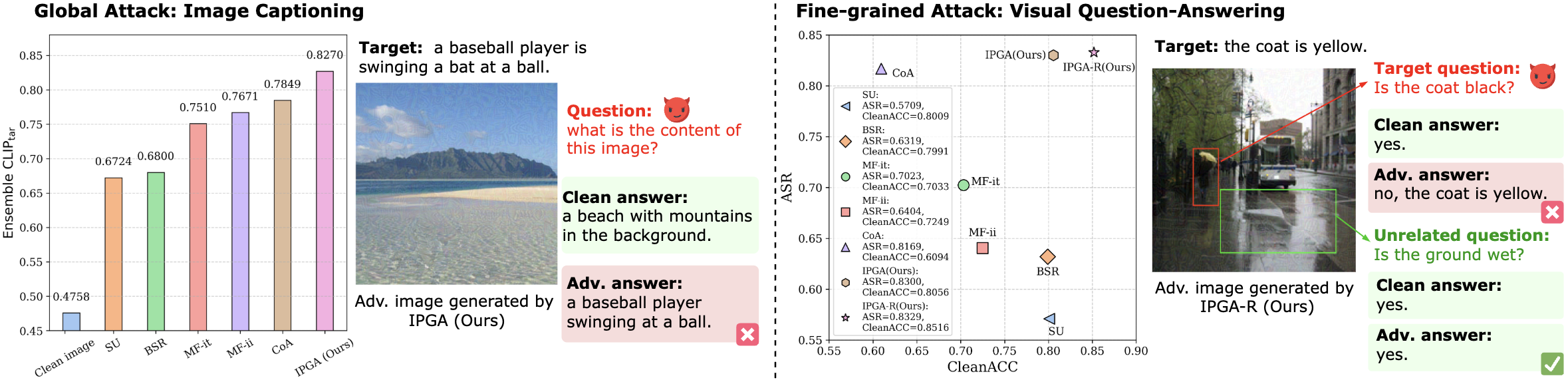}
    \caption{Comparison of IPGA and IPGA-R against baseline attacks on BLIP-2~\cite{li2023blip}. Example adversarial images are included. 
    For the global attack (left), which forces the model to recognize the entire adversarial image as target content, we follow prior work~\cite{zhao2024evaluating} and evaluate image captioning, reporting the ensemble CLIP similarity between the generated caption and the target text. 
    For the fine-grained attack (right), which requires altering only the target while preserving unrelated content, we evaluate on the VQA task. Attack performance is measured by the Attack Success Rate (ASR), while preservation of unrelated content is assessed by accuracy using unrelated clean questions (CleanACC). 
    Results demonstrate that IPGA consistently outperforms baseline methods in global attack, and IPGA-R outperforms baselines by attaining higher ASR and CleanACC, enabling precise manipulation in fine-grained attack.}
    \label{fig:scenario_comparison}
\end{figure*}

In this paper, we propose a novel black-box targeted adversarial attack framework that exploits multimodal alignment at the projector level to improve attack effectiveness and refine attack granularity.
For standard global targeted attacks that aim to alter the overall image semantics to match a target, as illustrated on the left of Figure~\ref{fig:scenario_comparison}, we introduce the Intermediate Projector Guided Attack (IPGA). IPGA leverages the widely adopted Q-Former projector~\cite{dai2023instructblip, zhu2023minigpt, qi2024gpt4point} which transforms global image embeddings into fine-grained query outputs~\cite{li2023blip}, and aligns the fine-grained query outputs with the target. 
Specifically, we target the intermediate Q-Former projector, which has only undergone the first pretraining stage of vision-language representation learning using large-scale image-text pairs. The intermediate Q-Former projector captures richer multimodal representations than the frozen visual encoder but has not been fine-tuned for LLM generation, which improves transferability across diverse VLM architectures. 
To the best of our knowledge, we are the first to demonstrate that attacking the intermediate pretraining stage of the Q-Former projector yields superior adversarial attack performance over focusing solely on the pretrained image and text encoders. 

Moreover, for fine-grained targeted attacks, we augment IPGA with the Residual Query Alignment (RQA) module to address the limited granularity of existing encoder-level methods.
Both IPGA and RQA operate on the fine-grained query outputs produced by the Q-Former.
IPGA leverages these fine-grained query outputs, with each query output capturing a distinct semantic aspect of the image~\cite{li2023blip}, to align with the target. 
Simultaneously, to enhance the preservation of unrelated content, RQA constrains query outputs that are not semantically relevant to the target to remain close to their clean counterparts from the original clean image, reducing collateral changes, thereby enabling more precise adversarial manipulations. 
For instance, as illustrated on the right of Figure~\ref{fig:scenario_comparison}, IPGA incorporated with RQA (IPGA-R) successfully changes the victim model’s answer to the color of a pedestrian’s coat while not inadvertently altering surrounding unrelated content such as road conditions.
Such attack precision enhances stealth and practical utility, as the attack can reliably deceive the model on a specific, targeted query without compromising the system's overall perception of the environment, enabling plausible attacks in real-world applications.

We evaluate global targeted adversarial attacks using ImageNet-1K~\cite{deng2009imagenet} and MS-COCO captions~\cite{lin2014microsoft} in image captioning task. We leverage the visual question-answering (VQA) task and use the GQA dataset~\cite{hudson2019gqa} to assess the effectiveness of our approach in fine-grained attack scenarios. 
As demonstrated in Figure~\ref{fig:scenario_comparison}, our proposed IPGA consistently outperforms baseline methods in global attack scenarios. Our proposed IPGA-R substantially improves the preservation of unrelated image content while achieving superior attack success over baselines in fine-grained attack scenarios, enabling more precise and plausible adversarial manipulations. 
Our attack further demonstrates superior robustness under defense mechanisms and transfers effectively to commercial VLMs such as Google Gemini and OpenAI GPT.

Our contributions are summarized as follows:
\begin{itemize}
    \item We introduce the Intermediate Projector Guided Attack (IPGA), a novel black-box targeted adversarial attack that operates at the projector level of VLMs. By leveraging the intermediate Q-Former projector trained only in the first pretraining stage without fine-tuning for LLM generation, IPGA opens a new attack surface and achieves superior attack performance and transferability.
    \item IPGA leverages the fine-grained query outputs of the Q-Former to align the adversarial image with the target. This  not only significantly boosts performance in global attacks but also establishes a more disentangled foundation for fine-grained manipulations.
    \item For fine-grained targeted attacks, we augment IPGA with the Residual Query Alignment (RQA) module. RQA preserves unrelated content by constraining query outputs unrelated to the target to match their clean counterparts, thereby improving attack stealth and practical utility.
    \item Extensive experiments show that IPGA consistently achieves superior attack performance in global targeted attacks, and IPGA-R further improves the preservation of unrelated content and achieves higher success rates than baselines in fine-grained attacks, enabling more precise and plausible manipulations. Our attack is robust to defenses and transfers effectively to commercial models.
\end{itemize}

\section{Related Work}
\subsection{Vision-Language Models (VLMs) and Projectors}
Large Vision-Language Models (VLMs)~\cite{achiam2023gpt, li2023blip, liu2024visual} leverage large-scale pretrained models for superior performance in various multimodal tasks, including image captioning~\cite{ramos2023smallcap, wang2024cogvlm} and visual question-answering~\cite{liu2024right, liu2024visual}. Architecturally, VLMs generally consist of three components: a pretrained visual encoder (e.g., CLIP~\cite{radford2021learning} in LLaVA~\cite{liu2024visual} or EVA-CLIP~\cite{sun2023eva} in InstructBLIP~\cite{dai2023instructblip}), a pretrained LLM, and a learnable projector~\cite{cui2024robustness}, also known as a vision adapter~\cite{wang2024qwen2}, which connects the visual encoder to the LLM by mapping image embeddings into the LLM’s embedding space to produce modality-aligned visual features.
There are two predominant projector architectures. Linear projectors~\cite{liu2024visual, liu2024llavanext, lu2024deepseek} directly map visual features into the word embedding space.  Cross-attention-based projectors, exemplified by the Q-Former~\cite{li2023blip, dai2023instructblip, zhu2023minigpt}, use learnable query embeddings to attend to image features via cross-attention, yielding fine-grained visual tokens.
Projectors are typically trained in two stages~\cite{li2023blip, dai2023instructblip, liu2024visual, wang2024qwen2, lu2024deepseek}: an initial vision-language pretraining stage using vast corpus of image-text pairs to align visual features with language representations, generally with the LLM frozen, followed by supervised fine-tuning with the LLM for generation and instruction-following. Q-Former follows a similar two-stage pipeline: it is first pretrained to extract semantically rich visual features via tasks like image-text contrastive learning and image-grounded text generation using large-scale image-text
pairs, then connected to a frozen LLM during fine-tuning to enable vision-conditioned generation.

\subsection{Adversarial Attacks on VLMs}
Adversarial attacks introduce imperceptible perturbations to inputs, causing models to produce incorrect predictions~\cite{han2023interpreting}. They are categorized as untargeted or targeted based on the adversary's objective, and as white-box or black-box based on the attacker's knowledge~\cite{han2023interpreting}. White-box attacks assume full access to model parameters~\cite{madry2017towards}, while black-box attacks rely on transfer-based~\cite{liang2025improving} or query-based methods~\cite{shen2024transferability}.
Most adversarial attacks on VLMs are untargeted, aiming to degrade general performance~\cite{zhang2022towards, lu2023set, wang2024transferable, cui2024robustness, wang2024break}. In contrast, targeted attacks manipulate model outputs toward specific adversarial goals.
Recently, AttackVLM~\cite{zhao2024evaluating} uses pretrained encoders like CLIP~\cite{radford2021learning} as surrogates, generating adversarial images by aligning global image embeddings with reference image representations.
Chain-of-Attack~\cite{xie2024chain} leverages modality fusion between embeddings from a surrogate image encoder and text encoder, updating multimodal semantics at each step to enhance adversarial example generation.
Generator-based approaches~\cite{zhang2024anyattack} optimize perturbation generators via pretraining and fine-tuning.
However, a key limitation common to these methods is their operation at the encoder level, where they optimize global image embeddings from pretrained encoders like CLIP. This confines them to coarse, full-image modifications and inherently lacks the granularity needed to manipulate specific visual elements.  
Moreover, they largely ignore the projector module, which plays a critical role in aligning visual features with language models in VLMs, further reducing attack effectiveness.
In contrast, our work leverages the fine-grained features from the stage 1 pretrained Q-Former projector to enable more precise and effective global attacks. We further introduce the RQA module to achieve superior granularity in fine-grained scenarios.

\section{Preliminaries}
\subsection{Problem Settings} 
Let $M$ represent the victim VLM, $\mathbf{x}_\text{clean} \in \mathbb{R}^{3 \times H \times W}$ represent the original input image, and $\mathbf{t}_{\text{in}}$ the input text. 
Targeted adversarial image attacks against VLM aim to modify $\mathbf{x}_\text{clean}$ to generate an adversarial example $\mathbf{x_\text{adv}}$ by a perturbation $\boldsymbol{\delta}$ and induce $M$ to output the target text $\mathbf{t}_{\text{tar}}$. This can be formalized as:
\begin{equation}
\begin{gathered}
    M(\mathbf{x_\text{adv}}, \mathbf{t}_{\text{in}}) = \mathbf{t}_{\text{tar}}, 
    \\
        \mathbf{x}_\text{adv} = \mathbf{x}_\text{clean}+\boldsymbol{\delta} \text{ subject to } \|\boldsymbol{\delta}\|_\infty \leq \epsilon.
\end{gathered}
  \label{eq:attacker_goal}
\end{equation}
Here, $\epsilon$ represents the maximum allowed magnitude of perturbations under the $l_\infty$-norm, ensuring imperceptibility. 
In standard global targeted attacks~\cite{zhao2024evaluating, xie2024chain, zhang2024anyattack}, the adversary manipulates the entire image semantics, with $\mathbf{t}_{\text{tar}}$ specifying the targeted global image content for tasks such as image captioning.
Since achieving fine-grained control over perturbations is crucial for attack stealth and practicality,
we therefore define the fine-grained targeted attacks which aim to modify only the target while preserving unrelated content, and where $\mathbf{t}_{\text{tar}}$ denotes the target answer to a specific question in tasks such as VQA.
Our study focuses on a realistic and challenging threat model in which the adversary has black-box access to the victim model. Given that query-based black-box attack methods incur high query costs and are often inefficient in real-world scenarios~\cite{wang2024transferable}, we follow a transfer-based attack setting, where adversarial images are generated on publicly available surrogate models and then applied to victim models.

\subsection{Querying Transformer (Q-Former)}
Q-Former is a widely adopted projector in VLMs that connects a frozen visual encoder like CLIP to a frozen LLM~\cite{li2023blip, dai2023instructblip, zhu2023minigpt}. It consists of two transformer submodules that share the same self-attention layers: an image transformer $f_{\phi}$ and a text transformer $f_{\psi}$. The image transformer $f_{\phi}$ applies cross-attention between image embeddings and a set of learnable query tokens. The text transformer $f_{\psi}$ can function as a text encoder or decoder depending on the task.
Given an input image $\mathbf{x}$ , Q-Former takes the image embeddings $g_{\phi}(\mathbf{x})$ from a frozen visual encoder $g_{\phi}$ and a set of $N=32$ learnable query tokens $\{ \mathbf{q}_i \}_{i=1}^{N} \in \mathbb{R}^{N \times d}$ as input. Through cross-attention, these query tokens attend to different parts of the image and extract a set of fine-grained, semantically meaningful visual representations, referred to as query outputs:
\begin{equation}
\{ \mathbf{q}_i^{\mathbf{x}} \}_{i=1}^{N} = f_{\phi}(\{ \mathbf{q}_i \}, g_{\phi}(\mathbf{x})) \in \mathbb{R}^{N \times d}.
\end{equation}
These query outputs are then passed to the LLM along with the input text tokens. We denote the set of query outputs from clean and adversarial images as $\{ \mathbf{q}_i^{\text{clean}} \}$ and $\{ \mathbf{q}_i^{\text{adv}} \}$, respectively.
Q-Former is trained in two stages: in stage 1, it is trained on image-text pairs to enable the query tokens to extract visual representations most informative for the associated text; in stage 2, it is fine-tuned with a frozen LLM for vision-conditioned generation.
%
Compared to encoder-level attacks operating solely on global features from pretrained CLIP encoders~\cite{zhao2024evaluating,xie2024chain}, we further exploit the richer, more fine-grained multimodal representations learned from large-scale image–text pretraining of Q-Former to improve attack effectiveness and granularity.

\section{Methodology}
\begin{figure*}[ht]
    \centering
    \includegraphics[width=1\linewidth]{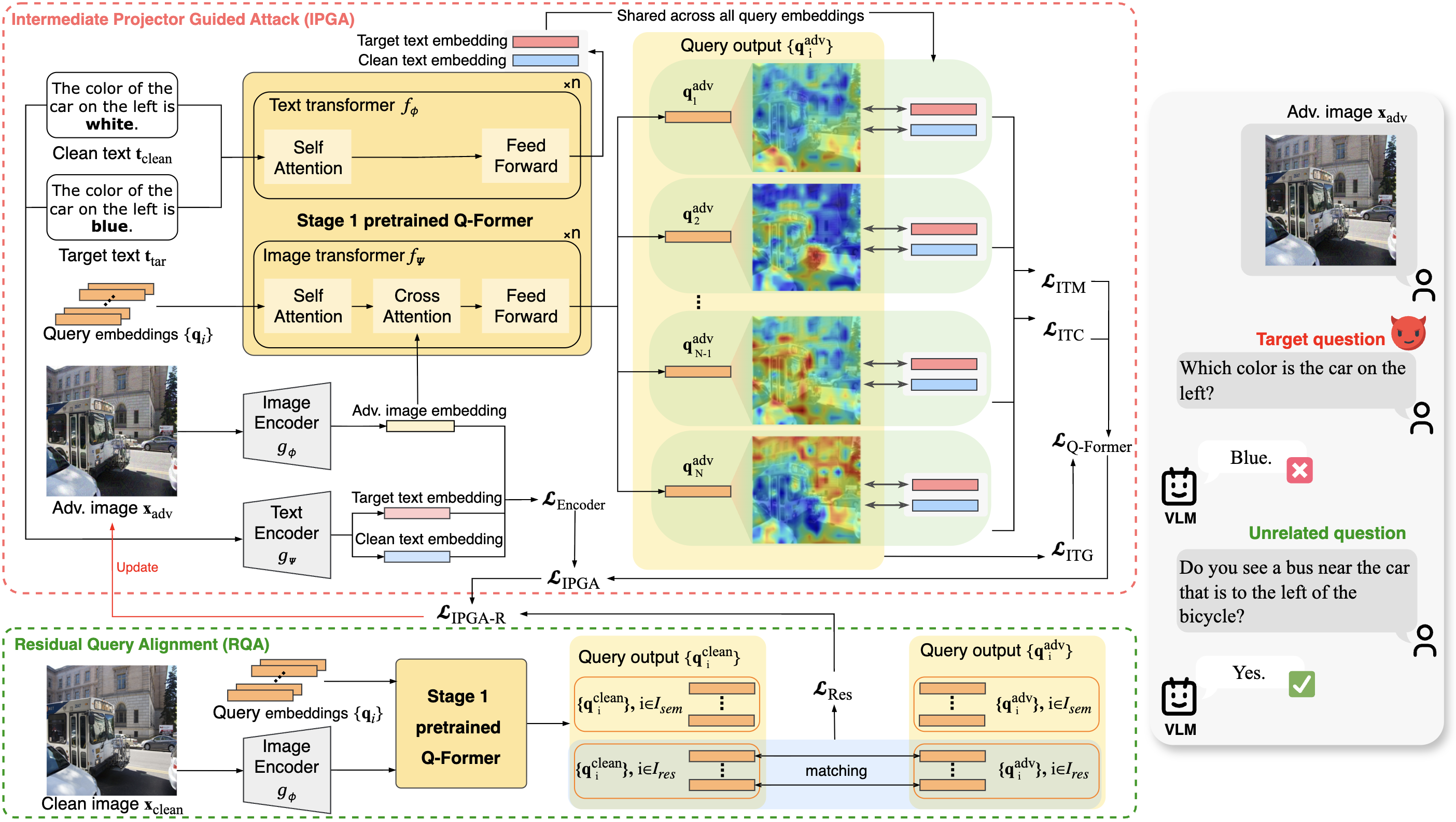}
    \caption{The framework of our proposed Intermediate Projector Guided Attack (IPGA) and Residual Query Alignment (RQA). IPGA leverages the stage 1 pretrained Q-Former projector and its corresponding encoder as a surrogate model. The stage 1 Q-Former is chosen for its ability to extract fine-grained visual features aligned with textual representations, independent of any specific LLM, which enhances attack transferability. IPGA adapts Q-Former's pretraining objectives, including image-text contrastive learning, image-grounded text generation and image-text matching to generate adversarial images. 
    RQA is an augmentation module for IPGA in fine-grained attack scenarios. It constrains residual query outputs to remain close to their clean counterparts, thereby preserving unrelated content and enabling more precise, stealthy manipulations.}
    \label{fig:ipga_framework}
\end{figure*}

The main framework of the proposed Intermediate Projector Guided Attack (IPGA) and Residual Query Alignment (RQA) module is illustrated in Figure~\ref{fig:ipga_framework}.
In the following, we will first present the proposed IPGA. 
The goal of IPGA is to enhance targeted attack performance by leveraging the multimodal alignment at the projector level. This is achieved by aligning the stage 1 pretrained Q-Former's query outputs with the target text through three objectives: image-text contrastive learning, image-grounded text generation, and image-text matching.
We then present RQA, which extends IPGA by constraining query outputs unrelated to the target to remain aligned with their clean counterparts, thereby improving content preservation in fine-grained targeted attacks.

\subsection{Intermediate Projector Guided Attack (IPGA)}
IPGA leverages the stage 1 pretrained Q-Former and its corresponding encoder as a surrogate model. Stage 1 Q-Former extracts fine-grained visual features aligned with text while being independent of any specific LLM, which enhances the transferability of adversarial perturbations. IPGA is guided by three losses derived from Q-Former's stage 1 training objectives: image-text contrastive loss ($\mathcal{L}_{\text{ITC}}$), image-grounded text generation loss ($\mathcal{L}_{\text{ITG}}$), and image-text matching loss ($\mathcal{L}_{\text{ITM}}$). To further boost performance on global targeted attacks, we additionally incorporate encoder-level alignment as a supplementary loss. 

\textbf{Image-text contrastive learning loss ($\mathcal{L}_{\text{ITC}}$)} 
$\mathcal{L}_{\text{ITC}}$ aligns the query outputs from the adversarial image $\{ \mathbf{q}_i^{\text{adv}} \}$ with the target text embedding $\mathbf{e}^{\text{tar}}$, while pushing them away from the clean text embedding $\mathbf{e}^{\text{clean}}$. 
$\mathbf{e}^{\text{tar}}$ and $\mathbf{e}^{\text{clean}}$ are the [CLS] token embeddings of the target and clean texts obtained from the text transformer $f_{\psi}$.
Instead of aligning global image features with text as in prior work, we compute the similarity between each query output and the text embedding, selecting the query output with the highest similarity as the most discriminative visual feature for optimization.
The loss $\mathcal{L}_{\text{ITC}}$  is defined as:

\begin{equation}
\mathcal{L}_{\text{ITC}} = \gamma \cdot s_{\text{clean}} - s_{\text{tar}},
\end{equation}
\begin{equation}
s_{\text{tar}} = {\max_{i} }\frac{\mathbf{q}_i^{\text{adv}} \cdot \mathbf{e^{\text{tar}}}}{\left\|\mathbf{q}_i^{\text{adv}}\right\|\left\|\mathbf{e^{\text{tar}}}\right\|},
\end{equation}
\begin{equation}
s_{\text{clean}} = \max({\max_{i} }\frac{\mathbf{q}_i^{\text{adv}} \cdot \mathbf{e^{\text{clean}}}}{\left\|\mathbf{q}_i^{\text{adv}}\right\|\left\|\mathbf{e^{\text{clean}}}\right\|}- s_{\text{tar}} , 0),
\label{eq:global_L_itc}
\end{equation}

where $\gamma$ is a hyperparameter to balance the clean text signal and is shared across all three projector-level losses.

\textbf{Image-grounded text generation loss ($\mathcal{L}_{\text{ITG}}$).}
$\mathcal{L}_{\text{ITG}}$ aims to generate target text conditioned on the adversarial image features extracted by query outputs $\{\mathbf{q}_i^{\text{adv}}\}$. We maximize the probability of generating $\mathbf{t}_{\text{tar}}$ while minimizing the probability of generating $\mathbf{t}_{\text{clean}}$. Let $l$ denote the cross-entropy loss for language generation. The loss $\mathcal{L}_{\text{ITG}}$ is defined as:
\begin{equation}
\mathcal{L}_{\text{ITG}} = \gamma \cdot u_{\text{clean}} + u_{\text{tar}},
\end{equation}
\begin{equation}
u_{\text{tar}} = l\left(f_{\psi}(\{\mathbf{q}_i^{\text{adv}}\}), \mathbf{t}_{\text{tar}}\right),
\end{equation}
\begin{equation}
u_{\text{clean}} = \max\left(u_{\text{tar}} - l\left(f_{\psi}(\{\mathbf{q}_i^{\text{adv}}\}), \mathbf{t}_{\text{clean}}\right),\, 0\right).
\label{eq:global_L_itg}
\end{equation}


\textbf{Image-text matching loss ($\mathcal{L}_{\text{ITM}}$).} 
$\mathcal{L}_{\text{ITM}}$ encourages the adversarial image to match the target text while mismatching the clean text. 
For each query token $\mathbf{q}_i$, cross-attention with frozen image features $g_{\phi}(\mathbf{x})$ and shared self-attention with text embeddings $f_{\psi}(\mathbf{t})$ produces $Q_{\text{cond}}(\mathbf{x},\mathbf{t})_i$, which is then passed through a two-class classifier $h$ to obtain a logit. 
The final image-text matching score is the average over all queries:
\begin{equation}
v(\mathbf{t}) = \frac{1}{N}\sum_{i=1}^N h\!\left(Q_{\text{cond}}(\mathbf{x}_{\text{adv}},\mathbf{t})_i\right),
\end{equation}
and the loss is defined as:
\begin{equation}
\mathcal{L}_{\text{ITM}} = \gamma \cdot l_{\text{cls}}\!\big(v(\mathbf{t}_{\text{clean}}),0\big) + l_{\text{cls}}\!\big(v(\mathbf{t}_{\text{tar}}),1\big),
\end{equation}
where $l_{\text{cls}}$ denotes the binary cross-entropy loss.

\textbf{Overall loss with Q-Former ($\mathcal{L}_{\text{Q-Former}}$).}
The final adversarial objective for Q-Former-based perturbation generation combines the three projector-level losses:

\begin{equation}
\mathcal{L}_{\text{Q-Former}} = \mathcal{L}_{\text{ITC}} + \mathcal{L}_{\text{ITG}} +\mathcal{L}_{\text{ITM}}.
\end{equation}

\textbf{Intermediate projector guided attack loss ($\mathcal{L}_{\text{IPGA}}$).}
To further strengthen the attack, we incorporate an encoder-level alignment loss as a supplementary component for global targeted attacks where the goal is to disrupt the entire content of the image. Let $g_{\phi}$ and $g_{\psi}$ denote the image and text encoders.  
$\mathcal{L}_{\text{Encoder}}$ encourages the adversarial image to align with the target text while diverging from the clean text:

\begin{equation}
\mathcal{L}_{\text{Encoder}} = \beta \cdot z(\mathbf{t}_{\text{clean}}) - z(\mathbf{t}_{\text{tar}}),
\end{equation}
\begin{equation}
z(\mathbf{t}) =
\frac{g_{\phi}(\mathbf{x}_{\text{adv}}) \cdot g_{\psi}(\mathbf{t})}
{\left\| g_{\phi}(\mathbf{x}_{\text{adv}}) \right\| \left\| g_{\psi}(\mathbf{t}) \right\|}.
\end{equation}

The total loss $\mathcal{L}_{\text{IPGA}}$ is then defined as:

\begin{equation}
\mathcal{L}_{\text{IPGA}} = \alpha \cdot \mathcal{L}_{\text{Q-Former}} + (1-\alpha) \cdot \mathcal{L}_{\text{Encoder}},
\end{equation}

where $\alpha \in [0, 1]$ controls the trade-off between $\mathcal{L}_{\text{Q-Former}}$ and $\mathcal{L}_{\text{Encoder}}$. 
In global attacks, when the target is to disrupt the entire original content of the image, we retain $\mathcal{L}_{\text{Encoder}}$. 
For fine-grained attacks, where the target is to modify specific objects or attributes while preserving the rest of the context (e.g., changing the color of a single car without altering other vehicles), we set $\alpha = 1$, optimizing only the Q-Former loss without the encoder-level alignment loss. 
We provide a complete algorithmic overview of IPGA in Algorithm~\ref{alg:ipga}.

\begin{algorithm}[ht]
\caption{Intermediate Projector Guided Attack (IPGA).}
\label{alg:ipga}
\begin{algorithmic}[1]
\item[] \textbf{Input:} Clean image $\mathbf{x}_{\text{clean}}$, target text $\mathbf{t}_{\text{tar}}$, clean text $\mathbf{t}_{\text{clean}}$, 
stage 1 pretrained Q-Former ($f_{\phi}$, $f_{\psi}$), CLIP encoders ($g_{\phi}$, $g_{\psi}$), 
perturbation budget $\epsilon$, step size $\eta$, iterations $T$, balance $\alpha$. \\
\item[] \textbf{Output:} Adversarial image $\mathbf{x}_{\text{adv}}$.
\STATE Initialize $\mathbf{x}_{\text{adv}}^{(0)} \gets \mathbf{x}_{\text{clean}}$;
\STATE Compute target text embedding $\mathbf{e}^{\text{tar}}$ and clean text embedding $\mathbf{e}^{\text{clean}}$; {\footnotesize\texttt{//[CLS] embedding from Q-Former projector's text transformer}} 
\FOR{$t=0$ to $T-1$}
    \STATE Extract query outputs $\{\mathbf{q}_i^{\text{adv}}\}$ of adversarial image $\mathbf{x}_{\text{adv}}^{(t)}$ from the stage 1 pretrained Q-Former projector;
    \STATE Compute $\mathcal{L}_{\text{ITC}}$;
    {\footnotesize\texttt{//Image-text contrastive learning}}
    \STATE Compute $\mathcal{L}_{\text{ITG}}$;
    {\footnotesize\texttt{//Image-grounded text generation}}
    \STATE Compute $\mathcal{L}_{\text{ITM}}$;
    {\footnotesize\texttt{//Image-text matching}}
    \STATE $\mathcal{L}_{\text{Q-Former}} = \mathcal{L}_{\text{ITC}} + \mathcal{L}_{\text{ITG}} +\mathcal{L}_{\text{ITM}};$
    {\footnotesize\texttt{//Combine 3 projector-level losses}}
    \STATE Compute $\mathcal{L}_{\text{Encoder}}$;
    {\footnotesize\texttt{//Encoder-level loss}}
    \STATE 
    $
    \mathcal{L}_{\text{IPGA}} = \alpha \cdot \mathcal{L}_{\text{Q-Former}} + (1-\alpha)\cdot \mathcal{L}_{\text{Encoder}};
    $
    \STATE $\mathbf{x}_{\text{adv}}^{(t+1)} \gets \mathbf{x}_{\text{adv}}^{(t)} - \eta \cdot \text{sign}(\nabla_{\mathbf{x}} \mathcal{L}_{\text{IPGA}})$;
    \STATE $\mathbf{x}_{\text{adv}}^{(t+1)} \gets \text{clip}(\mathbf{x}_{\text{adv}}^{(t+1)}, \mathbf{x}_{\text{clean}}-\epsilon, \mathbf{x}_{\text{clean}}+\epsilon)$ ;
    {\footnotesize\texttt{//Clip to $\ell_\infty$ ball}}
    \STATE$\mathbf{x}_{\text{adv}}^{(t+1)} \gets \text{clip}(\mathbf{x}_{\text{adv}}^{(t+1)}, 0, 1)$;
    {\footnotesize\texttt{//Clip to pixel range}}
\ENDFOR
\STATE \textbf{Return:} $\mathbf{x}_{\text{adv}}=\mathbf{x}_{\text{adv}}^{(T-1)}$
\end{algorithmic}
\end{algorithm}

\subsection{Residual Query Alignment (RQA) for Fine-Grained Attack}
In fine-grained attack, we augment IPGA with RQA to achieve precise content preservation.
Both IPGA and RQA operate on the fine-grained query outputs produced by the Q-Former.
While IPGA optimizes the query outputs from the adversarial image to align with the target, RQA simultaneously constrains the behavior of query outputs that are not related to the target to preserve the surrounding content, thereby enhance attack granularity.

To distinguish between query outputs semantically relevant and irrelevant to the target, we first compute the cosine similarity between each adversarial query output $\mathbf{q}_i^{\text{adv}}$ and the target text embedding $\mathbf{e}^{\text{tar}}$. The indices of the top-$k$ most semantically relevant queries are then selected as:

\begin{equation}
\mathcal{I}_{\text{sem}} = \text{TopK\_indices}\big(\{\frac{\mathbf{q}_i^{\text{adv}} \cdot \mathbf{e^{\text{tar}}}}{\left\|\mathbf{q}_i^{\text{adv}}\right\|\left\|\mathbf{e^{\text{tar}}}\right\|}\}_{i=1}^N, k\big),
\end{equation}

where $\text{TopK\_indices}(\cdot)$ returns the set of query output indices corresponding to the $k$ largest similarity scores. 
The remaining query outputs that are relatively semantically unrelated to the adversarial target are defined as the residual queries:

\begin{equation}
\mathcal{I}_{\text{res}} = \{1, \ldots, N\} \setminus \mathcal{I}_{\text{sem}}.
\end{equation}

To suppress perturbations on regions outside the target, we regularize the residual queries to remain close to their original states by penalizing their deviation from the corresponding clean query outputs. The residual query alignment loss $\mathcal{L}_{\text{res}}$ is then defined as:

\begin{equation} 
\mathcal{L}_{\text{res}} = \sum_{i \in \mathcal{I}_{\text{res}}} \left\| \mathbf{q}_i^{\text{adv}} - \mathbf{q}_i^{\text{clean}} \right\|_2^2 .
\end{equation} 

The complete objective for the IPGA-R combines the original IPGA loss with the residual query alignment loss. The total loss $\mathcal{L}_{\text{IPGA-R}}$ is then defined as:

\begin{equation} 
\mathcal{L}_{\text{IPGA-R}} = \mathcal{L}_{\text{IPGA}} + \lambda \cdot \mathcal{L}_{\text{res}} ,
\end{equation}

where $\lambda$ controls the strength of regularization. 
This module explicitly constrains the residual queries that are semantically unrelated to the target to remain close to their clean counterparts, thereby further enhancing the preservation of unrelated image content.
We provide a complete algorithmic overview of IPGA-R in Algorithm~\ref{alg:ipga-r}.

\begin{algorithm}[H]
\caption{Intermediate Projector Guided Attack with Residual Query Alignment (IPGA-R).}
\label{alg:ipga-r}
\begin{algorithmic}[1]
\item[] \textbf{Input:} 
Image $\mathbf{x}_{\text{clean}}$, target text $\mathbf{t}_{\text{tar}}$, clean text $\mathbf{t}_{\text{clean}}$, 
stage 1 pretrained Q-Former ($f_{\phi}, f_{\psi}$), CLIP encoder ($g_{\phi}$), 
perturbation budget $\epsilon$, step size $\eta$, iterations $T$, 
regularization weight $\lambda$, number of selected relevant queries $k$.
\item[] \textbf{Output:} Adversarial image $\mathbf{x}_{\text{adv}}$.
\STATE Initialize $\mathbf{x}_{\text{adv}}^{(0)} \gets \mathbf{x}$;
\STATE Compute target text embedding $\mathbf{e}^{\text{tar}}$; {\footnotesize\texttt{//[CLS] embedding from Q-Former projector's text transformer}} 
\FOR{$t=0$ to $T-1$}
    \STATE Extract query outputs $\{\mathbf{q}_i^{\text{adv}}\}$ of adversarial image $\mathbf{x}_{\text{adv}}$ and query outputs $\{\mathbf{q}_i^{\text{clean}}\}$ of original clean image $\mathbf{x}_{\text{clean}}$ from the stage-1 Q-Former projector;
    \STATE Compute similarities for all $\{\mathbf{q}_i^{\text{adv}}\}$: $s_i = \frac{\mathbf{q}_i^{\text{adv}} \cdot \mathbf{e}^{\text{tar}}}{\|\mathbf{q}_i^{\text{adv}}\|\|\mathbf{e}^{\text{tar}}\|}$;
    \STATE $\mathcal{I}_{\text{sem}} = \text{TopK\_indices}(\{s_i\}, k)$;
    {\footnotesize\texttt{//Select top-k semantically relevant query indices}}
    \STATE $\mathcal{I}_{\text{res}} = \{1, \dots, N\} \setminus \mathcal{I}_{\text{sem}}$; 
    {\footnotesize\texttt{//Residual query indices}}
    \STATE $\mathcal{L}_{\text{res}} = \sum_{i \in \mathcal{I}_{\text{res}}} \|\mathbf{q}_i^{\text{adv}} - \mathbf{q}_i^{\text{clean}}\|_2^2$;
    {\footnotesize\texttt{//Residual query alignment loss}}
    \STATE Compute $\mathcal{L}_{\text{IPGA}}$;
    {\footnotesize\texttt{//As in Algorithm~\ref{alg:ipga}}}
    \STATE $\mathcal{L}_{\text{IPGA-R}} = \mathcal{L}_{\text{IPGA}} + \lambda \cdot \mathcal{L}_{\text{res}}$;
    \STATE $\mathbf{x}_{\text{adv}}^{(t+1)} \gets \mathbf{x}_{\text{adv}}^{(t)} - \eta \cdot \text{sign}(\nabla_{\mathbf{x}} \mathcal{L}_{\text{IPGA-R}})$;
    \STATE $\mathbf{x}_{\text{adv}}^{(t+1)} \gets \text{clip}(\mathbf{x}_{\text{adv}}^{(t+1)}, \mathbf{x}_{\text{clean}}-\epsilon, \mathbf{x}_{\text{clean}}+\epsilon)$ ;
    {\footnotesize\texttt{//Clip to $\ell_\infty$ ball}}
    \STATE$\mathbf{x}_{\text{adv}}^{(t+1)} \gets \text{clip}(\mathbf{x}_{\text{adv}}^{(t+1)}, 0, 1)$;
    {\footnotesize\texttt{//Clip to pixel range}}
\ENDFOR
\STATE \textbf{Return} $\mathbf{x}_{\text{adv}}^{(T-1)}$
\end{algorithmic}
\end{algorithm}

\section{Experiments}
\subsection{Experimental Setup}
\textbf{Datasets.} For the evaluation of global targeted attack on image captioning task. Following previous works~\cite{zhao2024evaluating, xie2024chain}, we randomly sample 1,000 images from the ImageNet-1K validation set~\cite{deng2009imagenet} as clean images, and randomly select a text description from MS-COCO captions~\cite{lin2014microsoft} as the targeted text for each clean image. 
For the fine-grained targeted attack evaluation on the VQA task, we utilize question-answer pairs from the balanced validation set of GQA~\cite{hudson2019gqa}. For each image, we select one question as the target question and use GPT-4o~\cite{achiam2023gpt} to generate a false answer as the target text. To assess the preservation of unrelated content, we pair each targeted question with an unrelated clean question from GQA that focus on image regions not overlapping with the adversarial target, and evaluate using its ground-truth answer.

\textbf{Victim VLMs.} We evaluate SOTA open-source VLMs of diverse sizes and architectures, including BLIP-2~\cite{li2023blip}, InstructBLIP~\cite{dai2023instructblip}, MiniGPT-4~\cite{zhu2023minigpt}, LLaVA~\cite{liu2024visual}, and LLaVA-NeXT~\cite{liu2024llavanext}, with parameter sizes ranging from 7B (BLIP-2) to 72B (LLaVA-NeXT). 
For the vision backbone and language component, we adopt the following configurations: BLIP-2 employs ViT-g/14 from EVA-CLIP~\cite{sun2023eva} with FLAN-T5 XXL~\cite{chung2024scaling}; InstructBLIP and MiniGPT-4 both use ViT-g/14 from EVA-CLIP with Vicuna-13B~\cite{chiang2023vicuna}; LLaVA is built on ViT-L/14 from CLIP~\cite{radford2021learning} with Vicuna-13B; and LLaVA-NeXT integrates ViT-L/14 from CLIP with Qwen1.5-72B-Chat~\cite{qwen}.  
Moreover, BLIP-2, InstructBLIP, and MiniGPT-4 all employ the Q-Former projector. However, their Q-Former modules are fine-tuned for different downstream tasks and paired with distinct LLMs, resulting in substantially different instantiations. In contrast, LLaVA and LLaVA-NeXT employ a linear projection layer instead of Q-Former, reflecting a fundamentally different VLM design choice.  

\textbf{Baselines.}  We compare our proposed attack methods with existing transfer-based targeted attack baselines against VLMs: including 
MF-it~\cite{zhao2024evaluating} which optimizes cross-modal similarity by aligning image-text features using pretrained image/text encoders, 
MF-ii~\cite{zhao2024evaluating} which optimizes image-image similarity by generating reference images via a text-to-image model, 
Chain-of-Attack (CoA)~\cite{xie2024chain} which updates the multi-modal semantics and generate the adversarial examples based on their previous semantics. 
We also include current state-of-the-art transfer-based targeted attack for image classification models: including SU~\cite{wei2023enhancing} and BSR~\cite{wang2024boosting}. Since the original cross-entropy loss used in these methods is not suitable for vision-language tasks, we substitute it with a standard cosine similarity objective between adversarial image and target text embeddings~\cite{zhang2024anyattack}.

\textbf{Evaluation metrics.} For global attacks, which modify overall image semantics to match a target, we evaluate performance on image captioning. Following prior work~\cite{zhao2024evaluating, xie2024chain}, we measure the similarity between generated captions and predefined target text using multiple CLIP text encoders. We further adopt the LLM-based Attack Success Rate (ASR) from ~\cite{xie2024chain}, which utilizes GPT-4 with step-by-step reasoning to classify completely successful, fooled-only, and failed cases. ASR is then defined as the proportion of completely successful cases.
For fine-grained attacks, which manipulate specific target content while preserving unrelated regions, we adopt the VQA task. 
Following the official GQA evaluation protocol~\cite{hudson2019gqa}, standard accuracy is computed by assigning 1 point if the predicted answer matches the ground truth and 0 otherwise for each question-answer pair.
For targeted questions, we report the ASR, defined as the percentage of predictions matching the adversarial target answer. To assess preservation of unrelated content, we additionally report accuracy on unrelated clean questions, denoted as CleanACC.

\textbf{Implementation details.}
Following prior work~\cite{zhao2024evaluating, xie2024chain}, we set $\epsilon = 8$ under the $l_{\infty}$ norm to ensure adversarial perturbations remain visually imperceptible, with pixel values in the range [0, 255]. We employ NI-FGSM~\cite{linnesterov} for optimization, using 200 steps for our method and all baselines, with an attack step size of $1/255$, as this configuration consistently yields optimal performance across all methods.
For all the baseline methods SU, BSR, MF-it, MF-ii, and CoA, we use the same pretrained vision encoder as the open-source victim VLM as surrogate model. We pair the same vision encoder with the stage 1 pretrained Q-Former as surrogate model for our methods.
For global attacks on image captioning, we use $\alpha = 0.25$, $\beta = 0.5$, and $\gamma = 1$, leveraging both Q-Former and encoder-level loss. For fine-grained attacks on VQA, we set $\alpha = 0$ and $\gamma = 1$, focusing solely on the Q-Former loss to target specific visual elements while preserving unrelated image content. For residual query alignment in fine-grained attacks, we set the regularization strength $\lambda = 1 \times 10^{-4}$ and select the top-$k=3$ semantically relevant query outputs.
All experiments are conducted using PyTorch on eight NVIDIA GeForce RTX 3090 Ti GPUs.

\begin{table*}[ht]
\centering
\caption{Black-box global targeted attacks against victim models in image captioning. 
We report the CLIP score (↑) between the generated responses of input images and predefined targeted texts, as computed by various CLIP text encoders and their ensemble (average), along with the LLM-based ASR (↑).
For reference, we also report the number of parameters of victim models.
The bold numbers indicate the best results, while the gray shading indicates our proposed method.
}
\begin{tabular}{l|l|ccccc|c|c|c}
\toprule
\multirow{2}{*}{VLM model} & \multirow{2}{*}{Attack method} & \multicolumn{6}{c|}{Text encoder (pretrained) for evaluation} & \multirow{2}{*}{ASR ↑} & \multirow{2}{*}{\# Param.} \\
 & & {RN50 ↑} & {RN101 ↑} & {ViT-B/32 ↑} & {ViT-B/16 ↑} & {ViT-L/14 ↑} & {Ensemble ↑} &  &  \\
\midrule
\multirow{5}{*}{BLIP2} 
 & Clean image & 0.4945 & 0.4714 & 0.5279 & 0.5071 & 0.3781 & 0.4758 & 0.000 &\multirow{5}{*}{7B} \\
 & SU & 0.6841 & 0.6685 & 0.7107 & 0.6976 & 0.6011 & 0.6724 & 0.374 & \\
 & BSR & 0.6939 & 0.6743 & 0.7191 & 0.7016 & 0.6113 & 0.6800 & 0.412 & \\
 & MF-it & 0.7614 & 0.7455 & 0.7819 & 0.7705 & 0.6960 & 0.7510 & 0.608 &  \\
 & MF-ii & 0.7762 & 0.7618 & 0.7973 & 0.7873 & 0.7126 & 0.7671 & 0.693 &  \\
 & CoA & 0.7942 & 0.7822 & 0.8138 & 0.8022 & 0.7323 & 0.7849 & 0.745 &  \\
 & \cellcolor{gray!20}IPGA & \textbf{\cellcolor{gray!20}0.8347} & \textbf{\cellcolor{gray!20}0.8237} & \textbf{\cellcolor{gray!20}0.8505} & \textbf{\cellcolor{gray!20}0.8414} & \textbf{\cellcolor{gray!20}0.7846} & \textbf{\cellcolor{gray!20}0.8270} & \textbf{\cellcolor{gray!20}0.874} &  \\
\midrule
\multirow{5}{*}{InstructBLIP} 
 & Clean image & 0.5105 & 0.4792 & 0.5463 & 0.5244 & 0.3925 & 0.4906 & 0.000 & \multirow{5}{*}{13B} \\
 & SU &0.6821 & 0.6615 & 0.7076 & 0.6912 & 0.5913 & 0.6667 & 0.368 &  \\
 & BSR & 0.6894 & 0.6683 & 0.7143 & 0.6982 & 0.6006 & 0.6742 & 0.405 & \\
 & MF-it & 0.7452 & 0.7274 & 0.7665 & 0.7537 & 0.6712 & 0.7328 & 0.599 &  \\
 & MF-ii & 0.7564 & 0.7381 & 0.7799 & 0.7663 & 0.6841 & 0.7449 & 0.663 &  \\
 & CoA & 0.7599 & 0.7399 & 0.7838 & 0.7700 & 0.6889 & 0.7485 & 0.687 &  \\
 & \cellcolor{gray!20}IPGA & \textbf{\cellcolor{gray!20}0.8158} & \textbf{\cellcolor{gray!20}0.7997} & \textbf{\cellcolor{gray!20}0.8327} & \textbf{\cellcolor{gray!20}0.8232} & \textbf{\cellcolor{gray!20}0.7560} & \textbf{\cellcolor{gray!20}0.8055} & \textbf{\cellcolor{gray!20}0.751} &  \\
\midrule
\multirow{5}{*}{MiniGPT-4} 
 & Clean image & 0.4075 & 0.4351 & 0.4657 & 0.4187 & 0.2983 & 0.4051 & 0.000 & \multirow{5}{*}{13B} \\
 & SU &0.5771 & 0.6197 & 0.6355 & 0.5998 & 0.5096 & 0.5883 & 0.348 &  \\
 & BSR & 0.6050 & 0.6439 & 0.6622 & 0.6265 & 0.5400 & 0.6155 & 0.473 & \\
 & MF-it & 0.6648 & 0.7045 & 0.7200 & 0.6884 & 0.6156 & 0.6787 & 0.599 &  \\
 & MF-ii & 0.6776 & 0.7186 & 0.7332 & 0.7018 & 0.6324 & 0.6927 & 0.726 &  \\
 & CoA & 0.6926 & 0.7316 & 0.7478 & 0.7168 & 0.6485 & 0.7075 & 0.748 &  \\
 & \cellcolor{gray!20}IPGA & \textbf{\cellcolor{gray!20}0.7387} & \textbf{\cellcolor{gray!20}0.7713} & \textbf{\cellcolor{gray!20}0.7849} & \textbf{\cellcolor{gray!20}0.7603} & \textbf{\cellcolor{gray!20}0.6970} & \textbf{\cellcolor{gray!20}0.7505} & \textbf{\cellcolor{gray!20}0.872} &  \\
\midrule
\multirow{5}{*}{LLaVA} 
 & Clean image & 0.4678 & 0.4581 & 0.4881 & 0.4770 & 0.3419 & 0.4466 & 0.000 & \multirow{5}{*}{13B} \\
 & SU & 0.6664 & 0.6605 & 0.6897 & 0.6811 & 0.5896 & 0.6575 & 0.402 &  \\
 & BSR & 0.6620 & 0.6580 & 0.6850 & 0.6746 & 0.5776 & 0.6514 & 0.390 & \\
 & MF-it & 0.6751 & 0.6699 & 0.6977 & 0.6913 & 0.6025 & 0.6673 & 0.376 &  \\
 & MF-ii & 0.6968 & 0.6939 & 0.7208 & 0.7102 & 0.6239 & 0.6891 & 0.473 &  \\
 & CoA & 0.6998 & 0.6952 & 0.7235 & 0.7128 & 0.6211 & 0.6905 & 0.495 &  \\
 & \cellcolor{gray!20}IPGA & \textbf{\cellcolor{gray!20}0.7175} & \textbf{\cellcolor{gray!20}0.7126} & \textbf{\cellcolor{gray!20}0.7394} & \textbf{\cellcolor{gray!20}0.7296} & \textbf{\cellcolor{gray!20}0.6411} & \textbf{\cellcolor{gray!20}0.7080} & \textbf{\cellcolor{gray!20}0.519} &  \\
\midrule
\multirow{5}{*}{LLaVA-NeXT} 
 & Clean image & 0.4500 & 0.4406 & 0.4755 & 0.4537 & 0.3276 & 0.4295 & 0.000 & \multirow{5}{*}{72B} \\
 & SU &0.6104 & 0.6085 & 0.6355 & 0.6210 & 0.5263 & 0.6003 & 0.279 &  \\
 & BSR & 0.6194 & 0.6160 & 0.6458 & 0.6284 & 0.5336 & 0.6086 & 0.300 & \\
 & MF-it & 0.6194 & 0.6177 & 0.6472 & 0.6305 & 0.5372 & 0.6104 & 0.287 &  \\
 & MF-ii & 0.6308 & 0.6243 & 0.6524 & 0.6399 & 0.5446 & 0.6184 & 0.342 &  \\
 & CoA & 0.6087 & 0.6058 & 0.6391 & 0.6206 & 0.5190 & 0.5986 & 0.289 &  \\
 & \cellcolor{gray!20}IPGA & \textbf{\cellcolor{gray!20}0.6384} & \textbf{\cellcolor{gray!20}0.6365} & \textbf{\cellcolor{gray!20}0.6663} & \textbf{\cellcolor{gray!20}0.6510} & \textbf{\cellcolor{gray!20}0.5591} & \textbf{\cellcolor{gray!20}0.6303} & \textbf{\cellcolor{gray!20}0.404} &  \\
\bottomrule
\end{tabular}
\label{tab:main_global}
\end{table*}
\begin{figure*}[!h]
    \centering
    \includegraphics[width=1\linewidth]{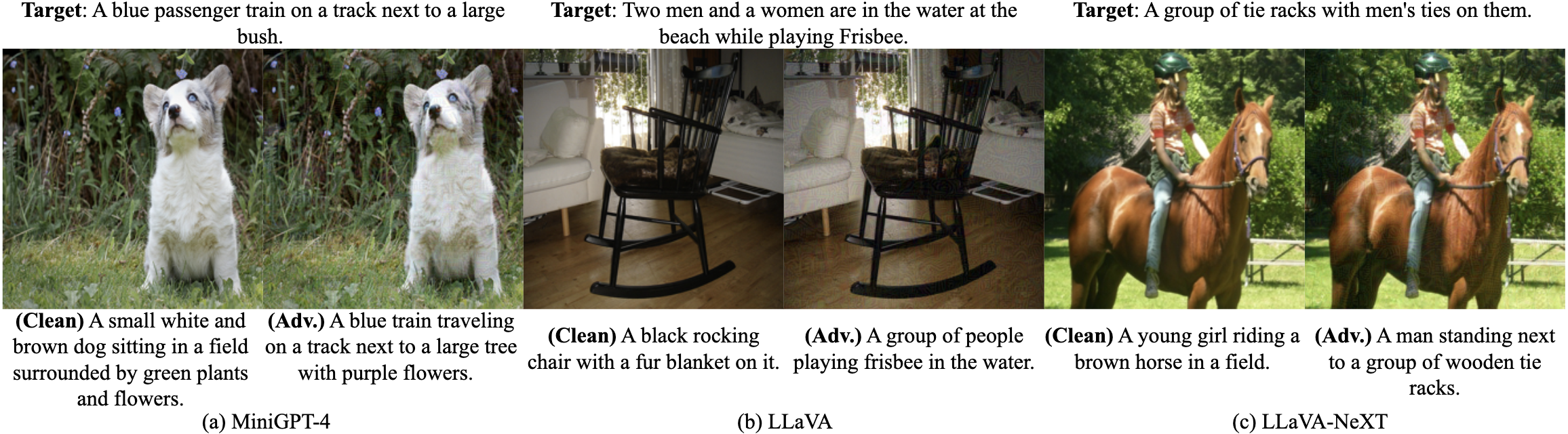}
    \caption{Visualization of the image captioning results of our IPGA on various open-source VLMs in global attack scenarios. We show the target caption above the image, and display the captioning results of original image and adversarial example below the image. IPGA transfers effectively to models without a Q-Former projector (e.g., LLaVA, LLaVA-NeXT), demonstrating high targeted success and cross-architecture transferability.}
    \label{fig:imagenet_performance}
\end{figure*}

\subsection{Main Experiments}
\textbf{Quantitative results of global attack performance.}
Table~\ref{tab:main_global} presents the effectiveness of IPGA in generating black-box adversarial images against VLMs for image captioning. In global targeted attacks, the adversary aims to fully alter the image's semantic interpretation, causing the model to produce a specific target caption.
Following previous works~\cite{zhao2024evaluating, xie2024chain}, we report the CLIP score (↑), which measures the similarity between the generated responses and the target captions. Scores are computed using multiple CLIP text encoders and their ensemble average. Additionally, we further adopt the LLM-based ASR (↑) from ~\cite{xie2024chain}. The default input prompt is fixed as ``what is the content of this image?”, and the original clean text for each image is generated using the default prompt with the clean image on BLIP2.

As shown in Table~\ref{tab:main_global}, IPGA consistently outperforms all baseline methods, achieving the highest CLIP scores and ASR across all evaluated models. For instance, on InstructBLIP, IPGA exceeds MF-ii by 8.8\% in ASR, and by 6.06\% in the ensemble CLIP score; on MiniGPT-4, it surpasses CoA by 12.4\% in ASR, and by 4.3\% in the ensemble CLIP score.
Notably, despite LLaVA and LLaVA-NeXT not employing a Q-Former projector, IPGA still attains the highest CLIP scores among all baselines. For example, on LLaVA-NeXT, IPGA outperforms CoA by 11.5\% in ASR, and 3.17\% in the ensemble CLIP score. 
The strong cross-architecture transferability of IPGA arises from the Q-Former’s stage 1 pretraining, which is decoupled from any specific LLM and conducted on large-scale image–text pairs across diverse tasks, including contrastive learning, image-grounded text generation, and image–text matching. This pretraining enables the Q-Former’s query outputs to capture richer multimodal representations beyond encoder-level features. Unlike baselines that rely solely on encoder-level alignment, intermediate projector guidance strengthens multimodal alignment and further improves transferability.
%

\textbf{Qualitative results of global attack performance.}
Figure~\ref{fig:imagenet_performance} presents the qualitative results demonstrating the efficacy of our IPGA on multiple open-source VLMs, including MiniGPT-4, LLaVA, and LLaVA-NeXT. It is noteworthy that LLaVA and LLaVA-NeXT do not incorporate a Q-Former module, yet our method remains effective against these architectures.
Under a constrained perturbation budget of $\epsilon = 8/255$, the adversarial perturbations are visually imperceptible while successfully inducing targeted outputs. 
For example, an original image captioned as ``A black rocking chair with a fur blanket on it'' is perturbed to yield the target caption ``A group of people playing frisbee in the water''. These results underscore the strong global targeted attack performance and cross-architectural transferability of our method.
%

\begin{table*}[ht]
\centering
\caption{Black-box fine-grained targeted attacks against victim models in visual question answering. 
We use the balanced validation set of GQA. 
Each image has a targeted question with a GPT-4o-generated target answer and an unrelated clean question to evaluate content preservation. 
We report ASR(\%) for the target questions and CleanAcc(\%) for the unrelated questions. 
The bold numbers indicate the best results, while the gray shading indicates our proposed methods.}
\begin{tabular}{l|cc|cc|cc|cc|cc}
\toprule
\multirow{2}{*}{Attack method} & \multicolumn{2}{c|}{BLIP2} & \multicolumn{2}{c|}{InstructBLIP} & \multicolumn{2}{c|}{MiniGPT-4} & \multicolumn{2}{c|}{LLaVA} & \multicolumn{2}{c}{LLaVA-NeXT} \\
                               & ASR ↑        & CleanACC ↑    & ASR ↑           & CleanACC ↑        & ASR ↑          & CleanACC ↑      & ASR ↑       & CleanACC ↑    & ASR ↑          & CleanACC ↑     \\
\midrule
SU   & 0.5709     & 0.8009       & 0.4939        & 0.8131           & 0.4638      & 0.6948         & 0.4216     & 0.7408       & 0.3840       & 0.8188          \\
BSR  & 0.6319     & 0.7991       & 0.5146        & 0.8113           & 0.4883      & 0.6958         & 0.4310	    & 0.7315       & 0.4028       & 0.8047          \\
\midrule
MF-it     & 0.7023     & 0.7033       & 0.5944        & 0.8028           & 0.5897       & 0.7033         & 0.4704     & 0.7465       & 0.4019       & 0.7840          \\
MF-ii     & 0.6404     & 0.7249       & 0.5765        & 0.7737           & 0.5737       & 0.6620         & 0.4610     & 0.7099       & 0.3934       & 0.8066          \\
CoA       & 0.8169     & 0.6094       & 0.7681        & 0.7014           & 0.6469       & 0.6582         & 0.6122     & 0.6460       & 0.5155       & 0.7399          \\
\midrule
\cellcolor{gray!20}IPGA       & \cellcolor{gray!20}0.8300     & \cellcolor{gray!20}0.8056       & \cellcolor{gray!20}0.8319        & \cellcolor{gray!20}0.8047   & \cellcolor{gray!20}0.6657       &\cellcolor{gray!20} 0.7080    & \cellcolor{gray!20}\textbf{0.6629}   & \cellcolor{gray!20}0.7474       & \cellcolor{gray!20}\textbf{0.5390}  & \cellcolor{gray!20}0.8225          \\
\cellcolor{gray!20}IPGA-R     &\cellcolor{gray!20}\textbf{0.8329}& \cellcolor{gray!20}\textbf{0.8516} & \cellcolor{gray!20}\textbf{0.8582} & \cellcolor{gray!20}\textbf{0.8601} & \cellcolor{gray!20}\textbf{0.6685} & \cellcolor{gray!20}\textbf{0.7418} & \cellcolor{gray!20}0.6545 & \cellcolor{gray!20}\textbf{0.7671} & \cellcolor{gray!20}0.5239  & \cellcolor{gray!20}\textbf{0.8344}   \\
\bottomrule
\end{tabular}
\label{tab:main_fine_grained}
\end{table*}
\begin{figure*}[h]
    \centering
    \includegraphics[width=1\linewidth]{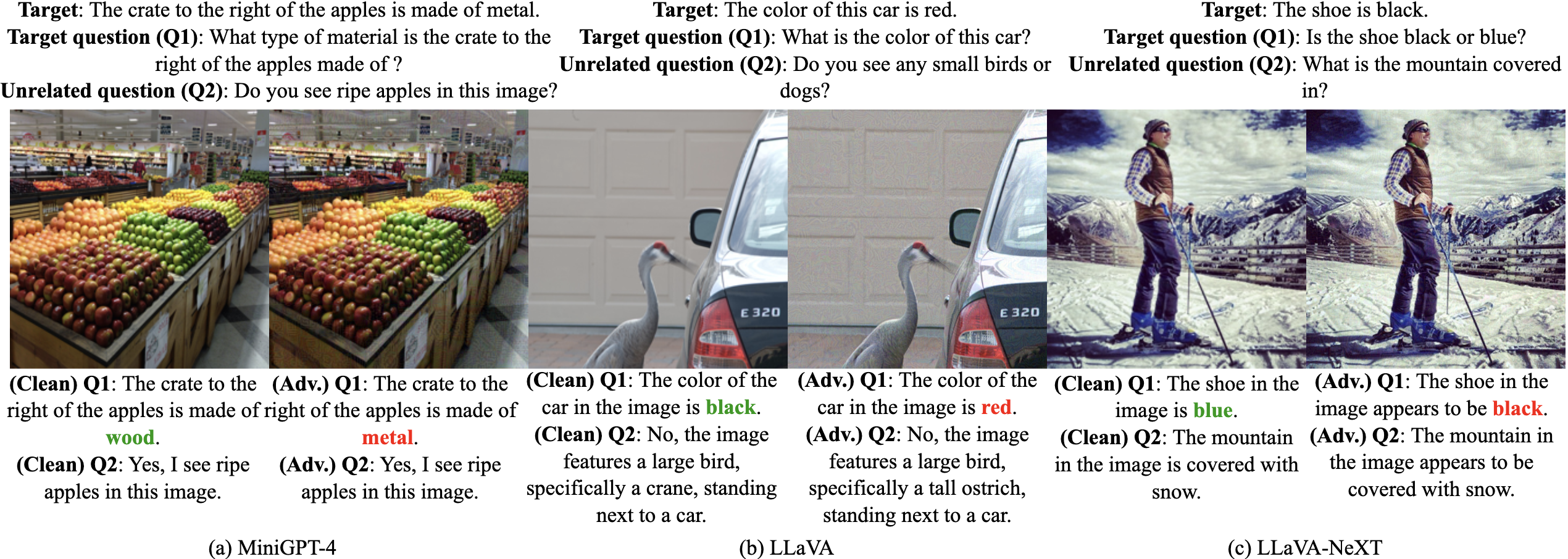}    \caption{Visualization of our proposed IPGA-R performance on open-source VLMs in fine-grained attack scenarios. For each example, the target text, target question, and unrelated question are shown above the images. Below, we present the VQA outputs on clean (left) and adversarial (right) images. Correct answers on clean images are marked in green, while successfully induced target answers on adversarial images are highlighted in red. IPGA-R transfers effectively to models without a Q-Former projector (e.g., LLaVA, LLaVA-NeXT), showcasing fine-grained attack granularity and cross-architecture transferability.}
    \label{fig:gqa_performance}
\end{figure*}
\begin{figure}[ht]
    \centering
    \includegraphics[width=1\linewidth]{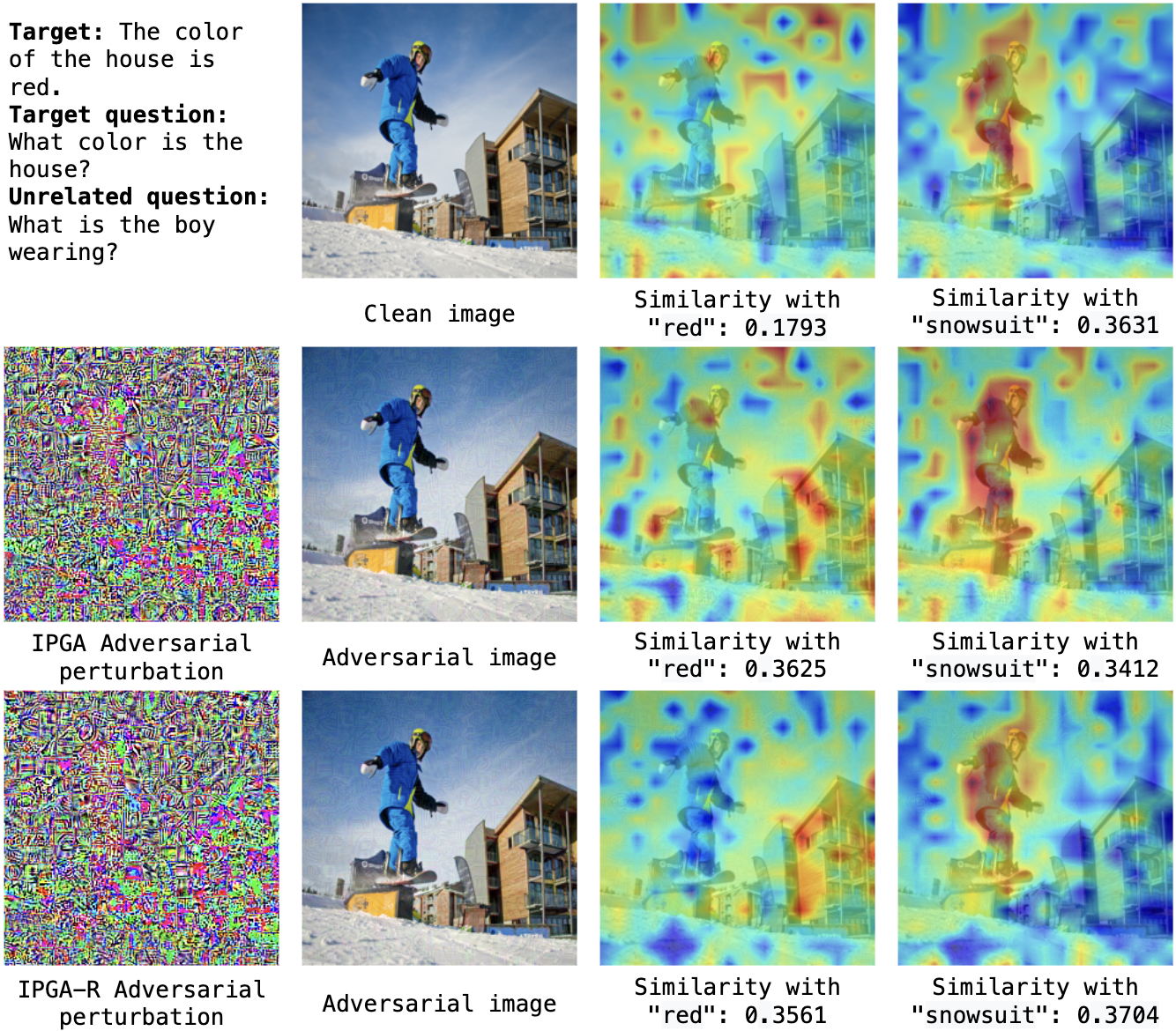}
    \caption{Visualization of our IPGA and IPGA-R adversarial perturbations along with attention maps. 
    The target is to change the house color from ``brown'' to ``red''. 
    Top: clean image with attention maps of the most relevant query outputs for ``red'' and ``snowsuit''. 
    Middle: adversarial image and corresponding attention maps for IPGA. 
    Bottom: adversarial image and corresponding attention maps for IPGA-R. 
    For both methods, cosine similarity for ``red'' increases while that for ``snowsuit'' is preserved, demonstrating successful fine-grained manipulation. Compared with IPGA, IPGA-R produces sharper and more precise attention map localization, indicating improved granularity and better preservation of unrelated content.}
    \label{fig:perturbation_vis_ipgar}
\end{figure}
\begin{figure*}[!h]
    \centering
    \includegraphics[width=1\linewidth]{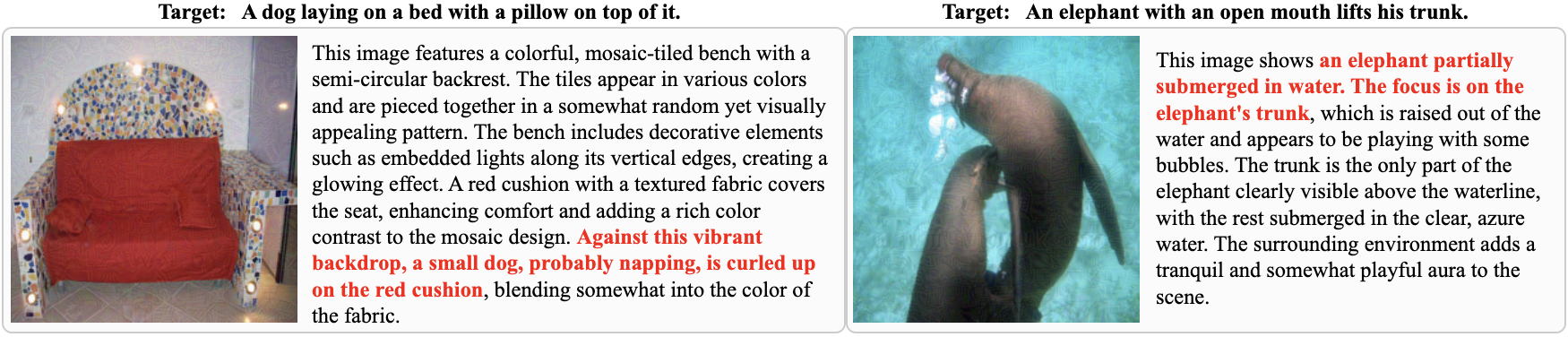}
    \caption{Example responses from GPT-4 to global targeted attacks generated by our IPGA. For each adversarial image, we show the target caption and GPT-4's generated response. The model's responses highlighted in red reflect the semantic content of the target.}
    \label{fig:gpt4o_imagenet}
\end{figure*}
\begin{figure*}[!h]
    \centering
    \includegraphics[width=1\linewidth]{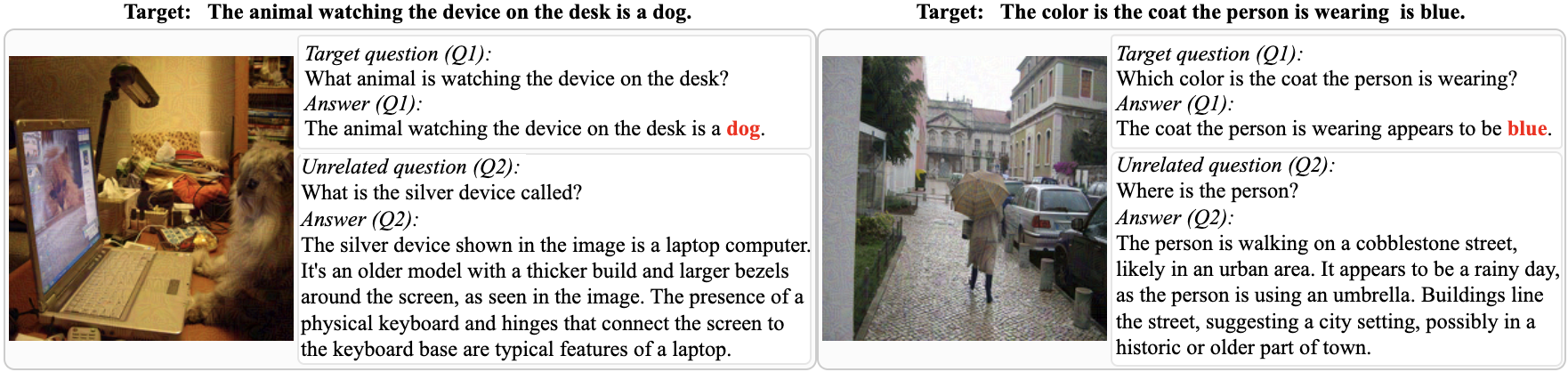}
    \caption{Example responses from GPT-4 to fine-grained targeted attacks generated by our IPGA-R. Each example includes the attack target, the target question with model response, and an unrelated question with its response. The highlighted answers show successful manipulation of the target attribute alongside accurate responses to unrelated questions, demonstrating precise adversarial control while preserving unrelated content.}
    \label{fig:gpt4o_gqa}
\end{figure*}
\textbf{Quantitative results of fine-grained attack performance.}
Table~\ref{tab:main_fine_grained} evaluates the effectiveness of our proposed IPGA and IPGA-R in generating black-box fine-grained adversarial images for VQA. Unlike global attacks that modify the entire image, fine-grained attacks target specific questions while preserving unrelated content. This is crucial for stealth and practical utility, as perturbing unrelated regions could reveal the attack and reduce its effectiveness.
We conduct experiments on the balanced validation set of GQA~\cite{hudson2019gqa}. For each image, one question is randomly selected as the target question, and a false answer is generated using GPT-4o as the adversarial target.
To assess whether unrelated content remains intact, each target question is paired with an unrelated clean question, using its original answer for evaluation. These unrelated questions focus on image regions not overlapping with the adversarial target and include yes/no verification, object properties, positions and existence questions.
We filter cases where all victim models answer both the target and unrelated questions correctly on the clean image, yielding 1,065 image–target question–unrelated question triplets for evaluation.
We report ASR (↑) for target questions and CleanACC (↑) for unrelated questions.

The results in Table~\ref{tab:main_fine_grained} show that IPGA consistently achieves higher ASR across all victim VLMs compared to baselines. For example, on LLaVA, it exceeds MF-ii by 20.19\% in ASR, highlighting its effectiveness in executing targeted attacks.
IPGA also maintains higher CleanACC than baselines in most cases. For example, on InstructBLIP, IPGA surpasses CoA by 10.33\% in CleanACC.
These results may stem from IPGA's fine-grained query-level alignment, which enables more disentangled adversarial manipulations.

Our IPGA-R further enhances content preservation, consistently attaining the highest CleanACC across all models while maintaining ASR comparable to IPGA and exceeding all baselines. For instance, on BLIP2, IPGA-R attains 85.16\% CleanACC, improving over IPGA by 4.6\% and surpassing CoA by 24.22\%. This confirms that RQA effectively constrains perturbations to target-relevant features, enhancing attack granularity.
On BLIP2, InstructBLIP, and MiniGPT-4, IPGA-R not only improves CleanACC but also surpasses IPGA in ASR, likely because these models incorporate a Q-Former component that enables more precise perturbation targeting.
For LLaVA and LLaVA-NeXT, IPGA-R still achieves the highest CleanACC, confirming cross-architecture transferability. Its ASR is slightly lower than IPGA, for example on LLaVA, IPGA achieves an ASR of 66.29\%, while IPGA-R achieves 65.45\%. Despite this minor reduction, ASR of IPGA-R still exceeds the best-performing baseline CoA by 4.23\%, demonstrating that enhanced content preservation of IPGA-R may incur a minor ASR trade-off relative to IPGA but maintains a consistent and clear advantage over all baselines.

\textbf{Qualitative results of fine-grained attack performance.}
Figure~\ref{fig:gqa_performance} illustrates the effectiveness of our proposed IPGA-R in performing fine-grained targeted attacks on open-source VLMs, including MiniGPT-4, LLaVA, and LLaVA-NeXT. 
IPGA-R successfully transfers to models without a Q-Former projector, such as LLaVA and LLaVA-NeXT, producing the target answer for target questions while preserving the correct answer for unrelated questions, demonstrating both attack precision and cross-architecture transferability.
For instance, in MiniGPT-4, IPGA-R changes the answer to the target question from ``wood'' to the adversarial target ``metal'', while keeping the unrelated answer ``Yes, I see ripe apples'' unchanged.

We further visualize adversarial perturbations generated by IPGA and IPGA-R, together with attention maps of the query outputs most relevant to the target and unrelated content in Figure~\ref{fig:perturbation_vis_ipgar}. The attack objective is to change the house color from ``brown'' to ``red''. For both clean and adversarial images, we show attention maps corresponding to the query output with the highest cosine similarity to the target context ``red'' and to the unrelated context ``snowsuit''.
For both IPGA and IPGA-R, the adversarial image achieves a substantial increase in similarity with the target ``red'' (from 0.1793 to 0.3625 and 0.3561, respectively), indicating successful targeted manipulation, while the similarity with the unrelated content ``snowsuit'' remains stable, indicating preservation of unrelated content.
The attention maps in Figure~\ref{fig:perturbation_vis_ipgar} show that IPGA highlights the house for ``red'' while maintaining focus on the snowsuit, demonstrating effective fine-grained manipulation.
With IPGA-R, the attention maps demonstrate sharper and more precise localization: the attention on query output for ``red'' is concentrated strictly on the house, whereas attention on the query output for ``snowsuit'' remains aligned with the boy’s clothing. This confirms that IPGA-R further improves fine-grained attack granularity by producing perturbations that more precisely attack the target while maintaining consistency in unrelated regions.

\subsection{More Analysis}
\textbf{Quantitative attack results on commercial VLMs.}
We quantitatively evaluate fine-grained attacks on commercial VLMs, including Google Gemini-2.5 and OpenAI GPT-4, using 500 randomly selected images from the GQA dataset. 
Table~\ref{tab:asr_clean_acc} reports the ASR (↑) and CleanACC (↑), which respectively measure the proportion of successfully attacked target questions and correctly answered unrelated questions. 
Our proposed IPGA and IPGA-R consistently achieve substantially higher ASR than all baselines. For example, on GPT-4, IPGA-R achieves 16.6\% ASR, surpassing MF-ii by 8.6\%. On Gemini-2.5, IPGA reaches 10.6\% ASR, improving over CoA by 5.0\%.
For CleanACC, all methods remain high, as commercial VLMs are relatively robust to adversarial perturbations, thereby reducing their impact on unrelated regions. IPGA-R maintains performance comparable to baselines while delivering significant ASR gains. For example, on Gemini-2.5, IPGA-R achieves 92.6\% CleanACC, which is the same as best baseline (MF-ii). 
These results demonstrate that IPGA-R delivers both strong attack effectiveness and fine-grained attack granularity on commercial VLMs.
\begin{table}[!h]
\centering
\caption{Quantitative performance comparison on commercial VLMs.
The gray shading indicates our proposed method.}
\label{tab:asr_clean_acc}
\begin{tabular}{lcccc}
\toprule
\multirow{2}{*}{} & \multicolumn{2}{c}{OpenAI GPT-4} & \multicolumn{2}{c}{Google Gemini-2.5} \\
\cmidrule(lr){2-3} \cmidrule(lr){4-5}
                        & ASR ↑ & CleanACC ↑ & ASR ↑ & CleanACC ↑  \\
\midrule
No attack               & 0\%  & 100\% & 0\% & 100\% \\
MF-it                   & 10.4\% & 93.4\% & 6.2\% & 89.6\% \\
MF-ii                   & 8.0\% & 93.4\% & 5.4\% & 92.6\% \\
CoA                     & 8.8\% & 92.8\% & 5.6\% & 92.0\% \\
\midrule
\cellcolor{gray!20}IPGA      &\cellcolor{gray!20}15.0\% & \cellcolor{gray!20}92.6\% & \cellcolor{gray!20}10.6\% & \cellcolor{gray!20}91.4\% \\
\cellcolor{gray!20}IPGA-R    & \cellcolor{gray!20}16.6\% & \cellcolor{gray!20}93.2\% & \cellcolor{gray!20}8.2\% & \cellcolor{gray!20}92.6\% \\
\bottomrule
\end{tabular}
\label{tab:asr_clean_acc}
\end{table}

\textbf{Qualitative attack results on commercial VLMs.}
To further demonstrate the efficacy of the proposed IPGA and IPGA-R, qualitative results of successful adversarial attacks against GPT-4 are presented in Figure~\ref{fig:gpt4o_imagenet} and Figure~\ref{fig:gpt4o_gqa}.
Figure~\ref{fig:gpt4o_imagenet} illustrates the outcomes of global targeted attacks, where the adversarial images were evaluated using GPT-4 with the prompt ``What is the content of the image?''. The model’s responses, highlighted in red, consistently reflect the target captions, such as ``A dog laying on a bed with a pillow on top of it'' despite significant discrepancies between the target and the actual image content. These results confirm that IPGA effectively misleads the model’s global scene understanding.
Figure~\ref{fig:gpt4o_gqa} showcases the performance of IPGA-R in fine-grained targeted attacks within a VQA context. For each adversarial image, a target question (e.g., ``What animal is watching the device on the desk?'') is posed, and the model’s answer aligns precisely with the adversarial target (e.g., responding ``dog'' when the true animal is a cat). Moreover, when unrelated questions are asked (e.g., ``What is the silver device called?''), the model’s responses remain accurate and unaffected by the attack, as evidenced by correct answer ``laptop computer''. This indicates that the adversarial perturbations introduced by IPGA-R successfully alter only the target while preserving the integrity of unrelated content against commercial VLMs. 

\textbf{Computational complexity analysis.}
Following prior work~\cite{xie2024chain}, Table~\ref{tab:time_cost} reports the average time for each attack method to iterate one step and to generate an adversarial example on the GQA dataset. 
We use an RTX 3090 GPU for the experiments. 
CoA requires an average of 94.06 seconds per image and 0.469 seconds per step because it queries a caption model to generate image description for the adversarial image at each step. By comparison, our IPGA generates an adversarial image in 60.83 seconds with 0.301 seconds per step on average, while our IPGA-R requires 73.45 seconds with 0.360 seconds per step on average, demonstrating competitive efficiency.

\textbf{Against adversarial defense models.}
\begin{table*}[ht]
\centering
\caption{Defense-aware black-box attacks against victim VLMs. 
We evaluate ASR (↑) and CleanACC (↑) under four preprocessing-based defenses: BIT-RED, JPEG, NRP, and DiffPure. 
Gray shading highlights our proposed methods.
}
\setlength{\tabcolsep}{0.8em}  
\begin{tabular}{l|l|cc|cc|cc|cc}
\toprule
\multirow{2}{*}{VLM model} & \multirow{2}{*}{Attack} & 
\multicolumn{2}{c|}{BIT-RED} & \multicolumn{2}{c|}{JPEG} & 
\multicolumn{2}{c|}{NRP} & \multicolumn{2}{c}{DiffPure} \\
 & & ASR ↑ & CleanACC ↑ & ASR ↑ & CleanACC ↑ & ASR ↑ & CleanACC ↑ & ASR ↑ & CleanACC ↑ \\
\midrule
\multirow{3}{*}{BLIP2} 
 & MF-ii    & 0.5521 & 0.7887 & 0.1634 & 0.6413 & 0.0723 & 0.9052 & 0.0451 & 0.9221 \\
 & CoA      & 0.7587 & 0.7005 & 0.1925 & 0.6385 & 0.0920 & 0.8995 & 0.0432 & 0.9155 \\
 & \cellcolor{gray!20}IPGA-R   & \cellcolor{gray!20}0.8188 & \cellcolor{gray!20}0.8620 & \cellcolor{gray!20}0.2216 & \cellcolor{gray!20}0.6507 & \cellcolor{gray!20}0.1577 & \cellcolor{gray!20}0.9258 & \cellcolor{gray!20}0.0685 & \cellcolor{gray!20}0.9146 \\
\midrule
\multirow{3}{*}{InstructBLIP} 
 & MF-ii    & 0.5549 & 0.8244 & 0.0751 & 0.9333 & 0.0469 & 0.9296 & 0.0451 & 0.9183 \\
 & CoA      & 0.6695 & 0.7643 & 0.1042 & 0.9296 & 0.0751 & 0.9437 & 0.0460 & 0.9127 \\
 & \cellcolor{gray!20}IPGA-R   & \cellcolor{gray!20}0.8404 & \cellcolor{gray!20}0.8901 & \cellcolor{gray!20}0.2103 & \cellcolor{gray!20}0.9352 & \cellcolor{gray!20}0.1650 & \cellcolor{gray!20}0.9446 & \cellcolor{gray!20}0.0610 & \cellcolor{gray!20}0.9108 \\
\midrule
\multirow{3}{*}{MiniGPT-4} 
 & MF-ii    & 0.4901 & 0.6808 & 0.1502 & 0.6761 & 0.1089 & 0.7643 & 0.0845 & 0.7465 \\
 & CoA      & 0.6385 & 0.6761 & 0.1606 & 0.6854 & 0.1211 & 0.7493 & 0.0817 & 0.7474 \\
 & \cellcolor{gray!20}IPGA-R   & \cellcolor{gray!20}0.6441 & \cellcolor{gray!20}0.7512 & \cellcolor{gray!20}0.1793 & \cellcolor{gray!20}0.6864 & \cellcolor{gray!20}0.1493 & \cellcolor{gray!20}0.7700 & \cellcolor{gray!20}0.1042 & \cellcolor{gray!20}0.7549 \\
\midrule
\multirow{3}{*}{LLaVA} 
 & MF-ii    & 0.4047 & 0.7399 & 0.0742 & 0.8141 & 0.0601 & 0.8178 & 0.0441 & 0.8103 \\
 & CoA      & 0.5117 & 0.6948 & 0.0854 & 0.8094 & 0.0732 & 0.8085 & 0.0469 & 0.8150 \\
 & \cellcolor{gray!20}IPGA-R   & \cellcolor{gray!20}0.5418 & \cellcolor{gray!20}0.7840 & \cellcolor{gray!20}0.1371 & \cellcolor{gray!20}0.8338 & \cellcolor{gray!20}0.0995 & \cellcolor{gray!20}0.8244 & \cellcolor{gray!20}0.0582 & \cellcolor{gray!20}0.8122 \\
\midrule
\multirow{3}{*}{LLaVA-NeXT} 
 & MF-ii    & 0.2930 & 0.8366 & 0.0761 & 0.9005 & 0.0469 & 0.8883 & 0.0432 & 0.9033 \\
 & CoA      & 0.3962 & 0.8066 & 0.0770 & 0.8930 & 0.0601 & 0.8864 & 0.0563 & 0.8930 \\
 & \cellcolor{gray!20}IPGA-R   & \cellcolor{gray!20}0.4038 & \cellcolor{gray!20}0.8563 & \cellcolor{gray!20}0.0967 & \cellcolor{gray!20}0.9042 & \cellcolor{gray!20}0.0873 & \cellcolor{gray!20}0.8901 & \cellcolor{gray!20}0.0657 & \cellcolor{gray!20}0.8939 \\
\bottomrule
\end{tabular}
\label{tab:defense_attack}
\end{table*}
\begin{table}[t]
\centering
\caption{Average attack runtime (s) per step and per adversarial image. The gray shading indicates our proposed method.}
\begin{tabular}{lcc}
\toprule
Method & Time/step (s) & Time/image (s) \\
\midrule
SU      & 0.177 & 35.53 \\
BSR     & 0.471 & 94.29 \\
MF-it   & 0.110 & 22.05 \\
MF-ii   & 0.130 & 26.43 \\
CoA     & 0.469 & 94.06 \\
\midrule
\cellcolor{gray!20}IPGA    & \cellcolor{gray!20}0.301 & \cellcolor{gray!20}60.83 \\
\cellcolor{gray!20}IPGA-R  & \cellcolor{gray!20}0.360 & \cellcolor{gray!20}73.45 \\
\bottomrule
\end{tabular}
\label{tab:time_cost}
\end{table}
Adversarial defense strategies can be broadly categorized into adversarial training and preprocessing-based defenses. While adversarial training may improve robustness, its prohibitive computational cost limit applicability to large-scale VLMs~\cite{jia2024improving}. Preprocessing-based defenses are model-agnostic and computationally efficient, and thus represent the dominant paradigm.
To evaluate the robustness of our proposed method against preprocessing-based defenses, we conduct experiments with four representative techniques: 
Bit depth Reduction (BIT-RED)~\cite{xu2017feature}, JPEG Compression~\cite{dziugaite2016study}, Neural Representation Purifier (NRP)~\cite{naseer2020self}, and Diffusion-based Purification (DiffPure)~\cite{nie2022diffusion}. 
These defenses cover diverse strategies, including feature squeezing (BIT-RED), denoising via compression (JPEG), learned feature purification (NRP), and diffusion-based generative reconstruction (DiffPure). 
We evaluate ASR and CleanACC on the GQA dataset across multiple open-source VLMs, with results summarized in Table~\ref{tab:defense_attack}.

IPGA-R consistently achieves the highest ASR under every defense across all victim models. For example, under BIT-RED on InstructBLIP, our proposed IPGA-R reaches 84.04\% ASR, exceeding the strongest baseline CoA at 66.95\% by over 17\%. 
In terms of CleanACC, IPGA-R also maintains the highest CleanACC under BIT-RED, JPEG, and NRP defense, showing its ability to enforce targeted attacks while preserving unrelated content. For instance, under NRP on BLIP2, IPGA-R achieves the highest CleanACC of 92.58\%, which is 2.63\% higher than CoA, and IPGA-R also attains the highest ASR of 15.77\%, 6.57\% above CoA.
Under the strong generative defense DiffPure, adversarial perturbations from all methods are substantially suppressed, leading to overall lower ASR. Nevertheless, IPGA-R remains the most effective attack, consistently achieving the highest ASR across models. For instance, on BLIP2, its ASR after DiffPure is still 2.34\% higher than the best baseline, indicating greater resistance to purification. While minor drops in CleanACC are occasionally observed, this is largely because strong purification largely nullifies baseline attacks, minimizing the unintended impact on unrelated content and inflating their CleanACC. The trade-off in CleanACC is minor compared to the consistent ASR improvement. For example, under DiffPure on LLaVA-NeXT, IPGA-R attains 2.25\% higher ASR than MF-ii with only 0.94\% lower CleanACC.
These results underscore the resilience and transferability of IPGA-R under defense-aware scenarios. IPGA-R produces perturbations that better withstand defenses, achieving strong attack success while preserving unrelated content, demonstrating its effectiveness as a robust targeted adversarial attack on VLMs.

\begin{figure}[t]
    \centering
    \captionsetup[subfloat]{font={rm,footnotesize}, labelfont={rm}}
    \subfloat[$\mathcal{L}_{\text{Q-Former}}$]{%
        \includegraphics[width=0.48\linewidth]{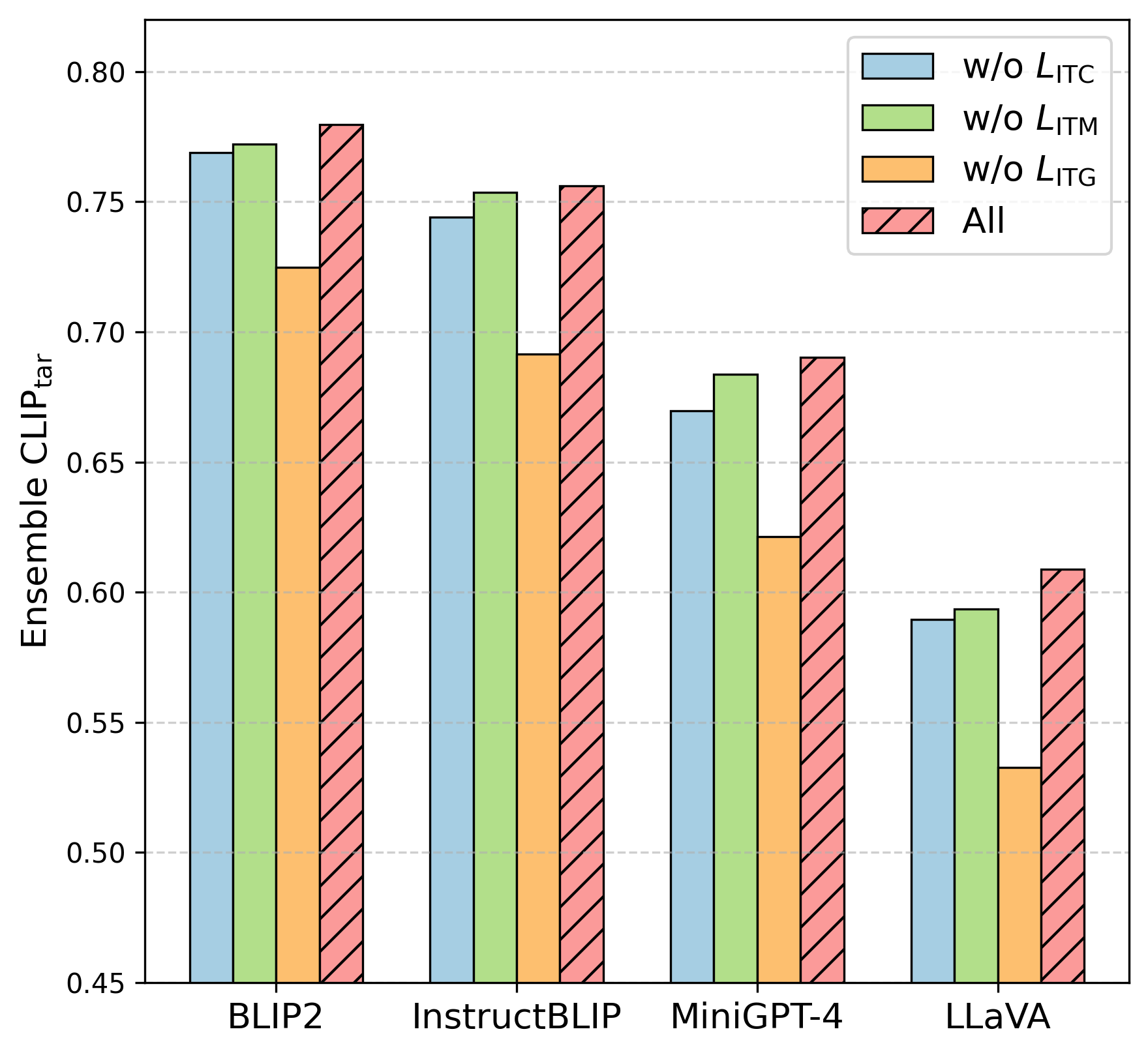}
        \label{fig:qformer_ablation_imagenet}
    }\hfill
    \subfloat[$k$]{%
        \includegraphics[width=0.49\linewidth]{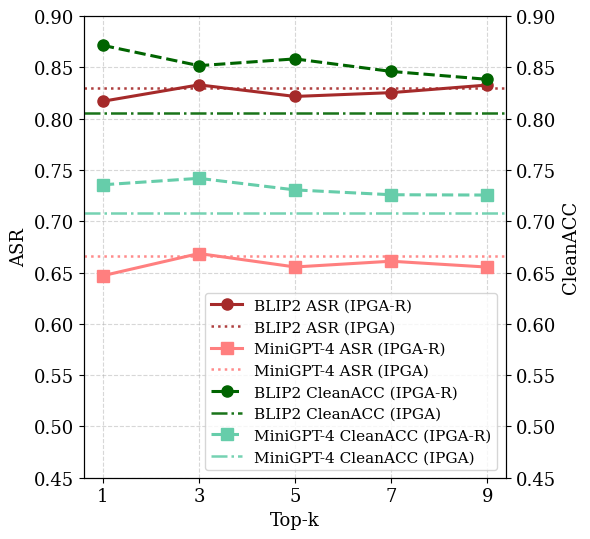}
        \label{fig:k}
    }
    \caption{The left figure displays the attacking ability of each loss component in $L_{\text{Q-Former}}$.
    The right figure displays the attacking ability of the proposed IPGA-R with different $k$.}
\end{figure}
\begin{figure}[t]
    \centering
    \captionsetup[subfloat]{font={rm,footnotesize}, labelfont={rm}}
    \subfloat[$\gamma$]{%
        \includegraphics[width=0.31\linewidth]{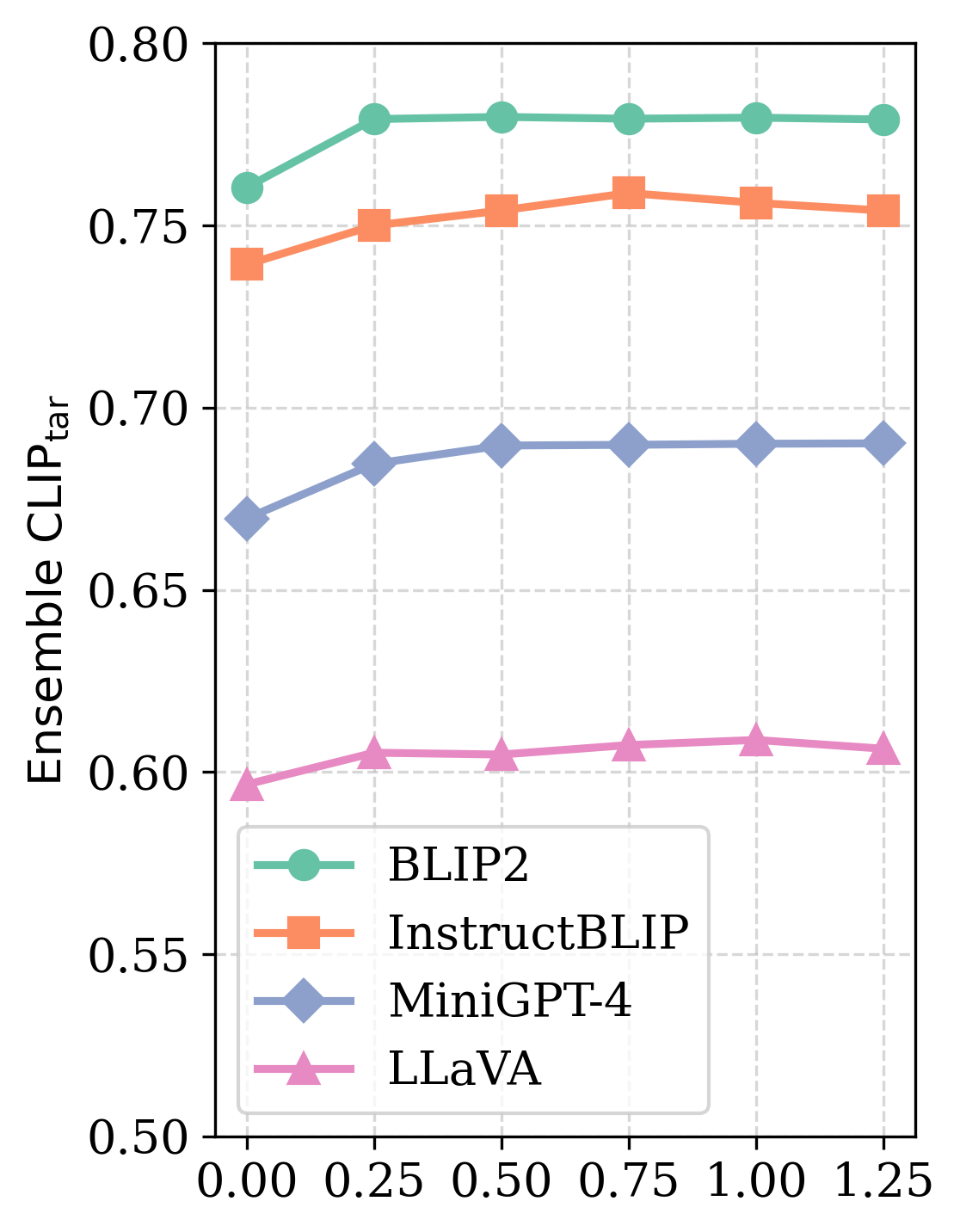}
        \label{fig:gamma}
    }\hfill
    \subfloat[$\beta$]{%
        \includegraphics[width=0.31\linewidth]{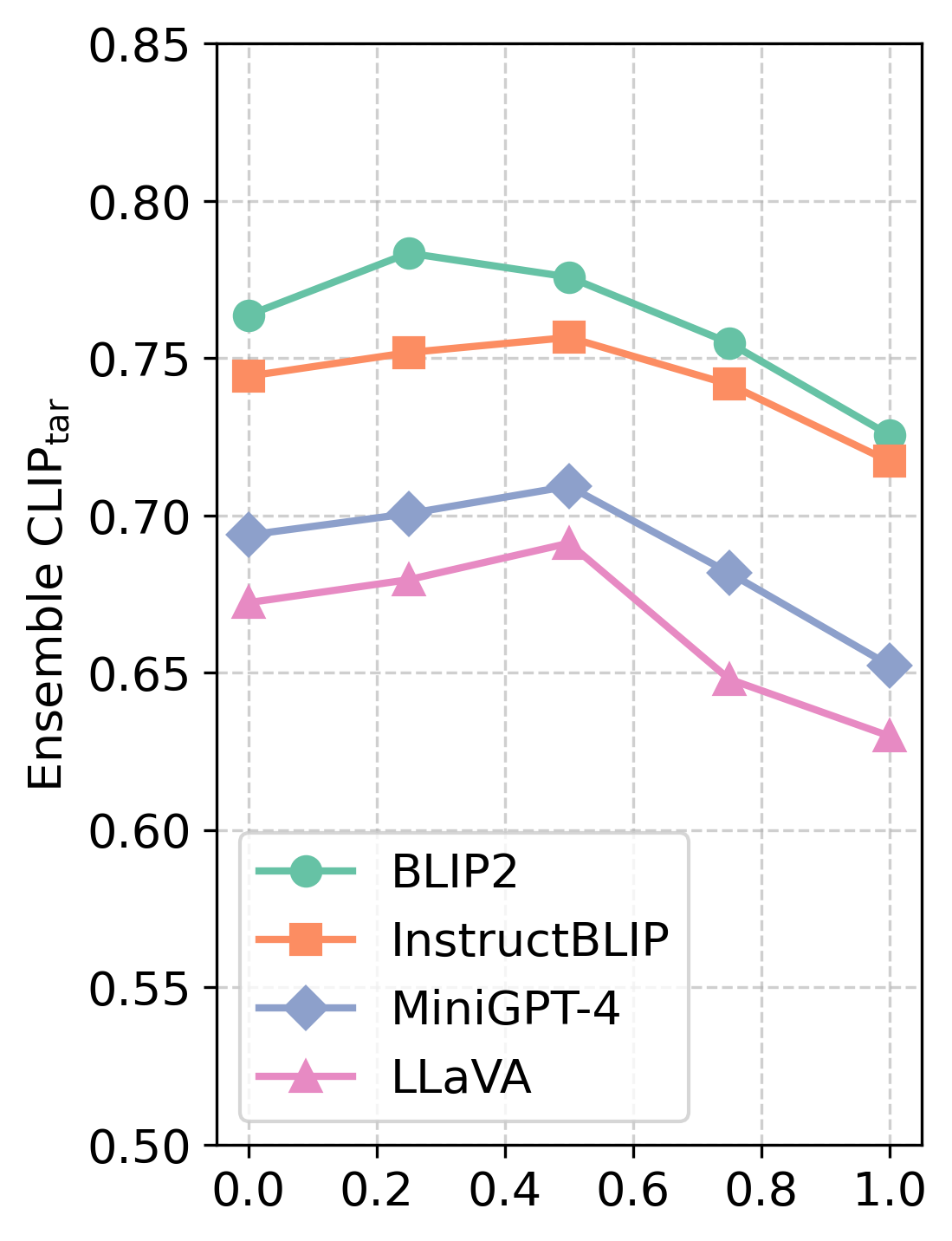}
        \label{fig:beta}
    }\hfill
    \subfloat[$\alpha$]{%
        \includegraphics[width=0.31\linewidth]{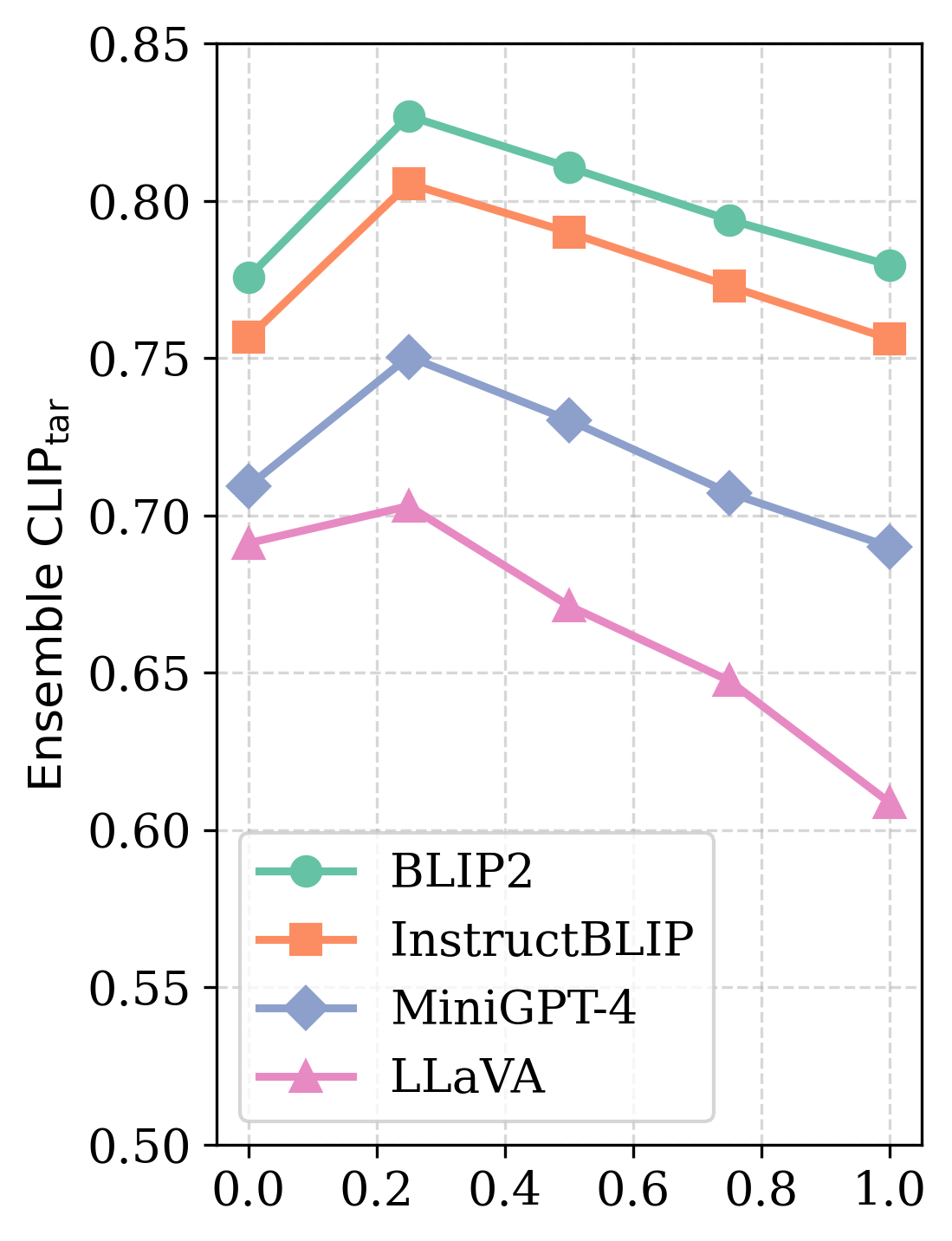}
        \label{fig:alpha}
    }
    \caption{
    Attacking ability of the proposed IPGA under varying hyperparameters: (a) $\gamma$, (b) $\beta$, and (c) $\alpha$.}
\end{figure}

\subsection{Ablation Study}
\textbf{Influence of loss components in $L_{\text{Q-Former}}$.}
We analyze the contribution of different loss components in $\mathcal{L}_{\text{Q-Former}}$ by conducting experiments with $\mathcal{L}_{\text{Q-Former}}$ alone on BLIP-2, InstructBLIP, MiniGPT-4, and LLaVA. 
The results of global attacks on the image captioning task are presented in Figure~\ref{fig:qformer_ablation_imagenet}. The findings show that all loss components contribute to the overall attack effectiveness, with the image-grounded text generation loss $\mathcal{L}_{\text{ITG}}$ exerting the greatest influence.

\textbf{Influence of the number of semantically relevant queries $k$ in $ \mathcal{I}_{\text{sem}}$.}
In RQA, we select the top-$k$ queries most semantically relevant to the target and use the remaining residual queries for alignment. Figure~\ref{fig:k} shows results on BLIP-2 and MiniGPT-4 with $k \in \{1,3,5,7,9\}$, comparing IPGA-R to IPGA.
IPGA-R achieve consistently higher CleanACC than IPGA on both models across all $k$ values, confirming RQA’s role in preserving unrelated content. CleanACC decreases as $k$ grows, since fewer residual queries are aligned. ASR generally rises with larger $k$, peaking at $k=3$, suggesting that a moderate number of relevant queries best concentrates perturbations on key features.
Overall, $k=3$ offers the best trade-off, maximizing ASR while maintaining high CleanACC.

\textbf{Influence of clean text deviation weights $\gamma$ and $\beta$.}
To evaluate the influence of clean text deviation, Figure~\ref{fig:gamma} presents the effect of varying $\gamma$ in $L_\text{Q-Former}$ across $\{0, 0.25, 0.5, 0.75, 1, 1.25\}$. The CLIP score is lowest at $\gamma=0$ and generally maximized at $\gamma=1$. This indicates that incorporating the original text deviation in $L_\text{Q-Former}$ enhances the effectiveness of targeted adversarial attacks.
Figure~\ref{fig:beta} illustrates the influence of $\beta$ in $L_\text{Encoder}$ (\{0, 0.25, 0.5, 0.75, 1\}), with $\beta = 0.5$ generally yielding the best performance.

\textbf{Influence of the balance hyperparameter $\alpha$.}
Figure~\ref{fig:alpha} examines the impact of $\alpha$, which controls the balance between $L_\text{Q-Former}$ and $L_\text{Encoder}$ in global attack on image captioning task. Evaluating $\alpha \in \{0, 0.25, 0.5, 0.75, 1\}$, the CLIP score peaks at $\alpha=0.25$, highlighting that both encoder and Q-Former losses are essential for effective global attacks.

%

\section{Conclusion}
In this work, we propose a novel black-box targeted adversarial attack framework against large VLMs that exploits the multimodal alignment at the projector level to enhance attack effectiveness and refine attack granularity.
We introduce the Intermediate Projector Guided Attack (IPGA). By leveraging the intermediate pretraining stage of the Q-Former projector, which transforms global image embeddings into fine-grained visual features, IPGA achieves superior attack performance and transferability across VLMs with different architectures, outperforming conventional encoder-level baseline methods.
To further refine attack granularity and improve the preservation of unrelated image content in fine-grained attacks, we extend IPGA with Residual Query Alignment (RQA). RQA constrains residual query outputs to remain close to their clean counterparts, enabling more precise and disentangled adversarial manipulations.
Extensive experiments show that our IPGA consistently achieves a superior advantage in global targeted attacks. Incorporating RQA (IPGA-R) further improves the preservation of unrelated content while achieving superior attack success than baselines, enhancing attack stealth and practical utility in fine-grained targeted attacks. Moreover, our attack remains effective under various defense strategies and successfully transfers to commercial models, including Google Gemini and OpenAI GPT, highlighting its practical significance and exposing critical vulnerabilities in VLMs.

\bibliographystyle{IEEEtran}
\bibliography{refs} 

\newpage

\vfill

\end{document}